\documentclass{article}

\usepackage{microtype}

\usepackage{times}
\usepackage{graphicx} 
\usepackage{subfigure} 
\usepackage{wrapfig}
\usepackage{natbib}

\usepackage{algorithm}
\usepackage{algorithmic}
\usepackage{siunitx}
\usepackage{booktabs}

\usepackage{hyperref}

\usepackage[accepted]{icml2017} 

\usepackage{enumitem}
\usepackage{Definitions}
\graphicspath{{./}}

\icmltitlerunning{Know-Evolve: Deep Temporal Reasoning for Dynamic Knowledge Graphs}

\begin{document} 

\twocolumn[
\icmltitle{Know-Evolve: Deep Temporal Reasoning for Dynamic Knowledge Graphs}




\begin{icmlauthorlist}
\icmlauthor{Rakshit Trivedi}{cc}
\icmlauthor{Hanjun Dai}{cc}
\icmlauthor{Yichen Wang}{cc}
\icmlauthor{Le Song}{cc}

\end{icmlauthorlist}
\icmlaffiliation{cc}{College of Computing, Georgia Institute of Technology}
\icmlcorrespondingauthor{Rakshit Trivedi}{rstrivedi@gatech.edu}
\icmlcorrespondingauthor{Le Song}{lsong@cc.gatech.edu}


\vskip 0.3in
]



\printAffiliationsAndNotice{} 

\begin{abstract} 
The availability of large scale event data with time stamps has given rise to \emph{dynamically evolving} knowledge graphs that contain temporal information for each edge. Reasoning over time in such dynamic knowledge graphs is not yet well understood. To this end, we present {\bf Know-Evolve}, a novel deep evolutionary knowledge network that learns non-linearly evolving entity representations over time. The occurrence of a fact (edge) is modeled as a multivariate point process whose intensity function is modulated by the score for that fact computed based on the learned entity embeddings. We demonstrate significantly improved performance over various relational learning approaches on two large scale real-world datasets. Further, our method effectively predicts occurrence or recurrence time of a fact which is novel compared to prior reasoning approaches in multi-relational setting.


%
\end{abstract} 
\vspace{0cm}
\section{Introduction}
\label{intro}

Reasoning is a key concept in artificial intelligence. A host of applications such as search engines, question-answering systems, conversational dialogue systems,  and social networks require  reasoning over underlying structured knowledge. Effective representation and learning over such knowledge has come to the fore as a very important task. In particular, Knowledge Graphs have gained much attention as an important model for studying complex multi-relational settings.
Traditionally, knowledge graphs are considered to be static snapshot of multi-relational data.
However, recent availability of large amount of event based interaction data that exhibits complex temporal dynamics in addition to its multi-relational nature has created the need for approaches that can characterize and reason over temporally evolving systems. For instance, GDELT  ~\cite{LeeSch13} and ICEWS ~\cite{BosLauObrSheStaWar17} are two popular event based data repository that contains evolving knowledge about entity interactions across the globe.

\begin{figure}[t]
\centering
\includegraphics[width = 0.3\textwidth]{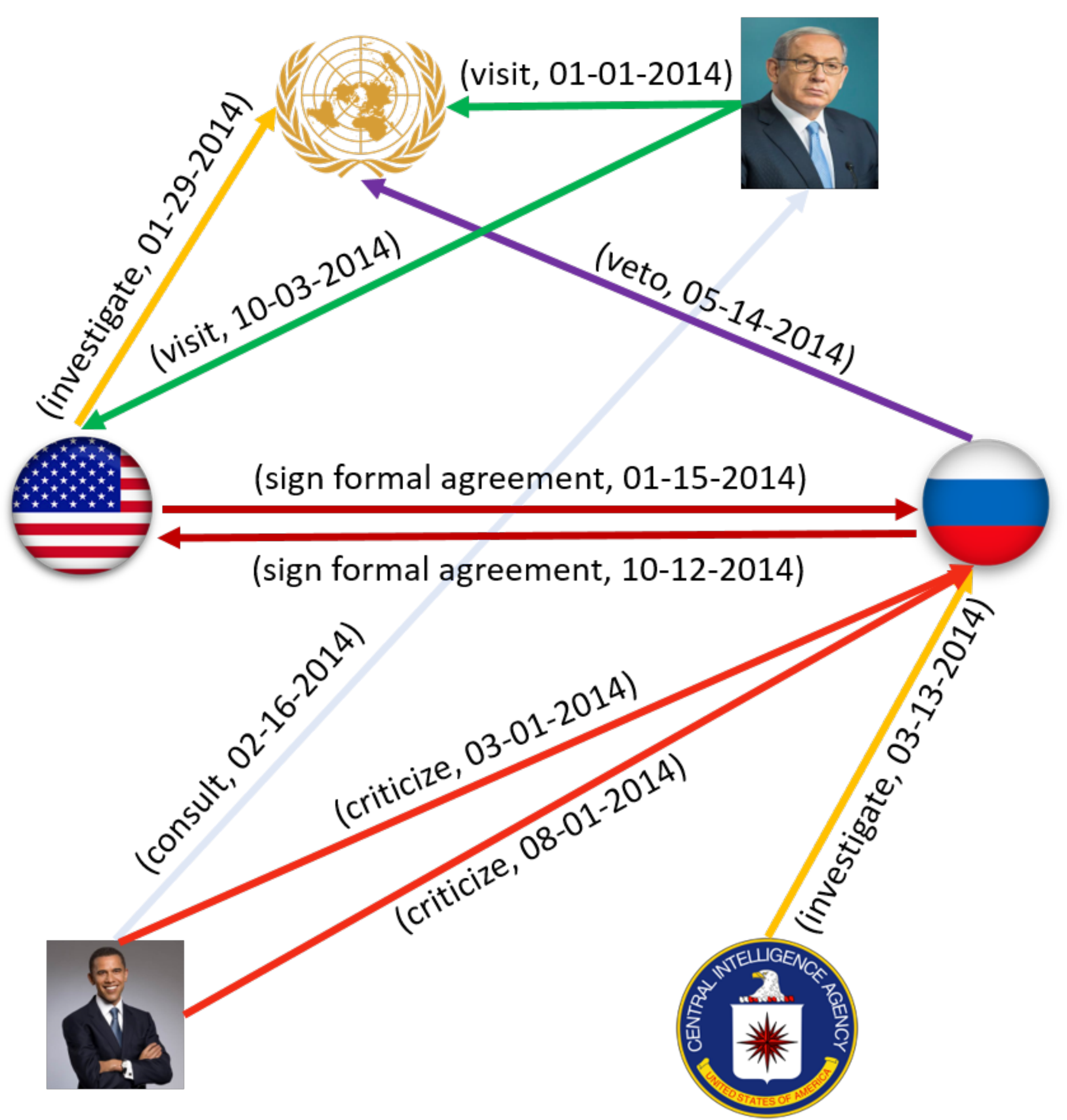}
\vspace{-2mm}
\caption{Sample temporal knowledge subgraph between persons, organizations and countries. }
\label{fig:kg}
\vspace{-0.5cm}
\end{figure}

Thus traditional knowledge graphs need to be augmented into \emph{Temporal Knowledge Graphs}, where facts occur, recur or evolve over time in these graphs, and each edge in the graphs have temporal information associated with it. Figure~\ref{fig:kg} shows a subgraph snapshot of such temporal knowledge graph. Static knowledge graphs suffer from incompleteness resulting in their limited reasoning ability. Most work on static graphs have therefore focussed on advancing entity-relationship representation learning to infer missing facts based on available knowledge. But these methods lack ability to use rich temporal dynamics available in underlying data represented by temporal knowledge graphs. 

Effectively capturing temporal dependencies across facts in addition to the relational (structural) dependencies can help improve the understanding on behavior of entities and how they contribute to generation of facts over time. For example, one can precisely answer questions like:
\begin{itemize}[leftmargin=*,nosep]
\item {\bf Object prediction}. (Who) will Donald Trump \underline{mention} next? 
\vspace{0.25cm}
\item {\bf Subject prediction}. (Which country) will \underline{provide material support} to US next month? 
\vspace{0.25cm}
\item {\bf Time prediction}. (When) will Bob \underline{visit} Burger King? 
\end{itemize}

\textit{"People (entities) change over time and so do relationships."} When two entities forge a relationship, the newly formed edge drives their preferences and behavior. This change is effected by combination of their own historical factors (\textbf{{temporal evolution}}) and their compatibility with the historical factors of the other entity (\textbf{{mutual evolution}}). 

For instance, if two countries have tense relationships, they are more likely to engage in conflicts. On the other hand, two countries forging an alliance are most likely to take confrontational stands against enemies of each other. Finally, time plays a vital role in this process. A country that was once peaceful may not have same characteristics 10 years in future due to various facts (events) that may occur during that period. Being able to capture this temporal and evolutionary effects can help us reason better about future relationship of an entity. We term this combined phenomenon of evolving entities  and their dynamically changing relationships  over time as \textbf{``knowledge evolution''}.
 
In this paper, we propose an elegant framework to model knowledge evolution and reason over complex non-linear interactions between entities in a multi-relational setting. The key idea of our work is to model the occurrence of a fact as multidimensional temporal point process whose conditional intensity function is modulated by the relationship score for that fact. The relationship score further depends on the dynamically evolving entity embeddings. Specifically, our work makes the following contributions:
\begin{itemize}[leftmargin=*,nosep]
\item We propose a novel deep learning architecture that evolves over time based on availability of new facts. The dynamically evolving network will ingest the incoming new facts, learn from them and update the embeddings of involved entities based on their recent relationships and temporal behavior.

\item Besides predicting the occurrence of a fact, our architecture has ability to predict time when the fact may potentially occur which is not possible by any prior relational learning approaches to the best of our knowledge. 

\item Our model supports \textit{Open World Assumption} as missing links are not considered to be false and may potentially occur in future. It further supports prediction over unseen entities due to its novel dynamic embedding process.
 
\item The large-scale experiments on two real world datasets show that our framework has consistently and significantly better performance for link prediction than state-of-arts that do not account for temporal and evolving non-linear dynamics.

\item Our work aims to introduce the use of powerful mathematical tool of temporal point process framework for temporal reasoning over dynamically evolving knowledge graphs. It has potential to open a new research direction in reasoning over time for various multi-relational settings with underlying spatio-temporal dynamics.
\end{itemize}

\vspace{-0.3cm}
\section{Preliminaries}
\subsection{Temporal Point Process}
A temporal point process~\cite{CoxLew2006} is a random process whose realization consists of a list of events localized in time, $\cbr{t_i}$ with $t_i \in \RR^+$. Equivalently, a given temporal point process can be re\-pre\-sen\-ted as a counting process, $N(t)$, which records the number of events before time $t$.

An important way to characterize temporal point processes is via the conditional intensity function $\lambda(t)$, a stochastic model for the time of the next event given all the previous events. Formally, $\lambda(t)\rd t$ is the conditional probability of observing an event in a small window $[t, t+\rd t)$ given the history $\Tcal(t):=\cbr{t_k|t_k<t}$ up to $t$, \ie,
\vspace{-0.2cm}
\begin{equation}
\begin{split}
\lambda(t)\rd t &:= \PP\cbr{\text{event in }[t, t+\rd t)|\Tcal(t)} \\
&= \EE[\rd N(t) | \Tcal(t)]
\end{split} 
\end{equation}
where one typically assumes that only one event can happen in a small window of size $\rd t$, \ie,~$\rd N(t) \in \cbr{0,1}$. 

From the survival analysis theory~\cite{AalBorGje08}, given the history $\Tcal=\cbr{t_1,\dotso, t_n}$, for any $t > t_n$, we characterize the conditional probability that no event happens during $[t_n, t)$ as $ S(t|\Tcal) = \exp\big(-\int_{t_n}^{t} \lambda(\tau) \, \rd\tau \big)$. Moreover, the conditional density that an event occurs at time $t$ is defined as :
\vspace{-0.3cm}
\begin{equation}
f(t) = \lambda(t)\, S(t)	
\label{eq:f}
\end{equation}
The functional form of the intensity $\lambda(t)$ is often designed to capture the phenomena of interests. Some Common forms include:
Poisson Process, Hawkes processes~\cite{Hawkes71}, Self-Correcting Process~\cite{IshWes79}, Power Law and Rayleigh Process.


{\bf Rayleigh Process} is a non-monotonic process and is well-adapted to modeling fads, where event likelihood drops rapidly after rising to a peak. Its intensity function is $\lambda(t) = \alpha \cdot (t)$, where $\alpha > 0$ is the weight parameter, and the log survival function is $ \log S(t|\alpha) = -\alpha \cdot (t)^ 2 / 2$. 

\vspace{-0.3cm}
\subsection{Temporal Knowledge Graph representation}

We define a \emph{Temporal Knowledge Graph (TKG)} as a multi-relational directed graph with timestamped edges between any pair of nodes. In a \emph{TKG}, each edge between two nodes represent an event in the real world and edge type (relationship) represent the corresponding event type. Further an edge may be available multiple times (recurrence). We do not allow duplicate edges and self-loops in graph. Hence, all recurrent edges will have different time points and every edge will have distinct subject and object entities.
 
Given $n_e$ entities and $n_r$ relationships, we extend traditional  triplet representation for knowledge graphs to introduce time dimension and represent each fact in \emph{TKG} as a quadruplet $(e^s, r, e^o, t)$,  where $e^s, e^o \in \{1, \ldots, n_e\}$, $e^s \ne e^o$, $r \in \{1, \ldots, n_r\}$, $t \in \RR^+$. It represents the creation of relationship edge $r$ between subject entity $e^s$, and object entity $e^o$ at time $t$. The complete TKG can therefore be represented as an $n_e \times n_e \times n_r \times \mathcal{T}$ - dimensional tensor where $\mathcal{T}$ is the total number of available time points.  
Consider a TKG comprising of $N$ edges and denote the globally ordered set of corresponding N observed events as $\mathcal{D} = \{(e^s, r, e^o, t)_n\}_{n=1}^N$, where $0~\le~t_1~\le~t_2~\ldots\le~T$.

\vspace{-0.3cm}
\section{Evolutionary Knowledge Network} 
\vspace{-0.1cm}

We present our unified knowledge evolution framework (Know-Evolve) for reasoning over temporal knowledge graphs. 
The reasoning power of Know-Evolve stems from the following three major components:
\vspace{-0.1cm}
\begin{enumerate}
\item A powerful mathematical tool of temporal point process that models occurrence of a fact.
\vspace{-0.1cm}
\item A bilinear relationship score that captures multi-relational interactions between entities and modulates the intensity function of above point process.
\vspace{-0.1cm}
\item  A novel deep recurrent network that learns non-linearly and mutually evolving latent representations of entities based on their interactions with other entities in multi-relational space over time. 
\end{enumerate}
\vspace{-0.4cm}
\subsection{Temporal Process}

Large scale temporal knowledge graphs exhibit highly heterogeneous temporal patterns of events between entities. Discrete epoch based methods to model such temporal behavior fail to capture the underlying intricate temporal dependencies. We therefore model time as a random variable and use temporal point process to model occurrence of fact. 

More concretely, given a set of observed events $\mathcal{O}$ corresponding to a \emph{TKG}, we construct a relationship-modulated  multidimensional point process to model occurrence of these events. We characterize this point process with the following conditional intensity function:  
\begin{equation}
\label{eq:intent}
	\lambda^{e^s,e^o}_r(t|\bar{t}) = f(g^{e^s,e^o}_r(\bar{t})) * (t-\bar{t}) 
\end{equation}
where $t > \bar{t}$, $t$ is the time of the current event and $\bar{t} = max(t^{e^s}-,t^{e^o}-)$ is the most recent time point when either subject or object entity was involved in an event before time $t$. Thus, $\lambda^{e^s,e^o}_r(t|\bar{t})$ represents intensity of event involving triplet $(e^s, r, e^j)$ at time $t$ given previous time point $\bar{t}$ when either $e^s$ or $e^o$ was involved in an event. This modulates the intensity of current event based on most recent activity on either entities' timeline and allows to capture scenarios like non-periodic events and previously unseen events. $f( \cdot ) = \exp( \cdot )$ ensures that intensity is positive and well defined. 

\vspace{-0.2cm}
\subsection{Relational Score Function}
\vspace{-0.1cm}

The first term in~(\ref{eq:intent}) modulates the intensity function by the relational compatibility  score between the involved entities in that specific relationship. Specifically, for an event $(e^s, r, e^o, t) \in \mathcal{D}$  occurring at time $t$, the score term $g^{e^s,e^o}_r$ is computed using a bilinear formulation as follows:
\begin{equation}
\label{eq:sc}
g^{e^s,e^o}_r(t) = \mathbf{v^{e^s}}(t-)^T \cdot \mathbf{R_r} \cdot \mathbf{v^{e^o}}(t-)
\end{equation}
where $\mathbf{v^{e^s}}$, $\mathbf{v^{e^s}} \in \RR^d$ represent latent feature embeddings of entities appearing in subject and object position respectively. $\mathbf{R_r} \in \RR^{d \times d}$ represents relationship weight matrix which attempts to capture interaction between two entities in the specific relationship space $r$. This matrix is unique for each relation in dataset and is learned during training. $t$ is time of current event and $t-$ represent time point just before time $t$. $\mathbf{v^{e^s}}(t-)$ and $\mathbf{v^{e^o}}(t-)$, therefore represent most recently updated vector embeddings of subject and object entities respectively before time $t$. As these entity embeddings evolve and update over time, $g^{e^s,e^o}_r(t)$ is able to capture cumulative knowledge learned about the entities over the history of events that have affected their embeddings. 
\vspace{-0.2cm}
\subsection{Dynamically Evolving Entity Representations}
\label{subsec:entity_embed}
\begin{figure*}[ht!]
\small
\centering
\includegraphics[width = 1\textwidth]{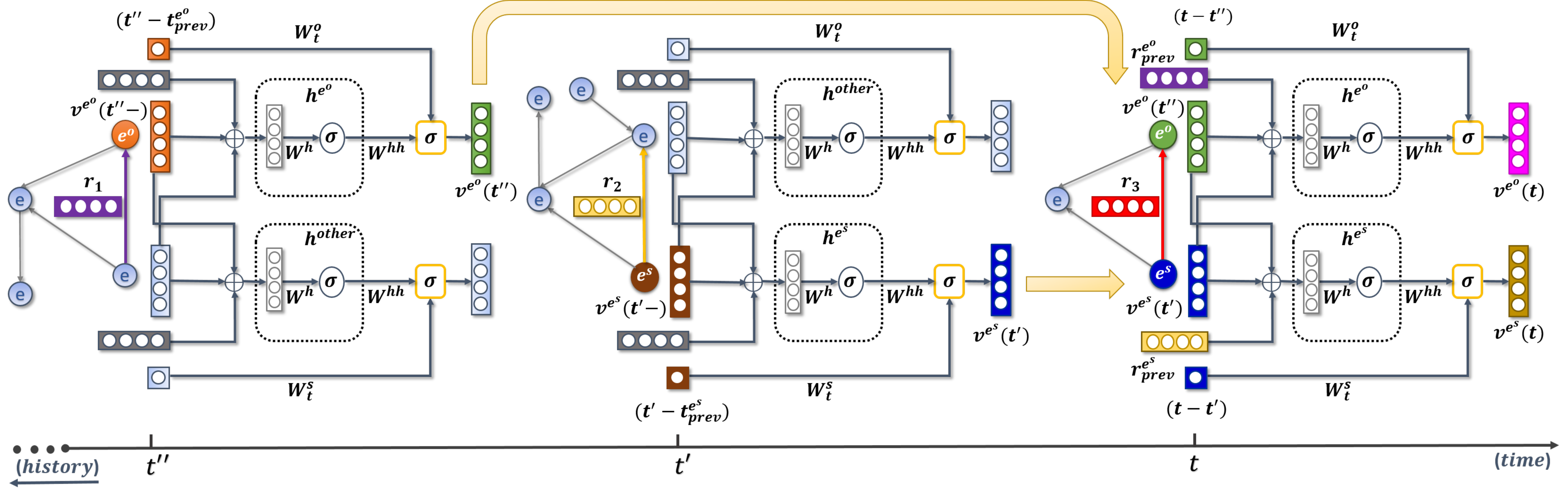}
\vspace{-0.7cm}
\caption{Realization of Evolutionary Knowledge Network Architecture over a timeline. Here $t''$, $t'$ and $t$ may or may not be consecutive time points. We focus on the event at time point $t$ and show how previous events affected the embeddings of entities involved in this event. From Eq.~(\ref{eq:sub}) and (\ref{eq:obj}), $t_{p-1} = t'$ and $t_{q-1} = t''$ respectively. $t_{prev}^{e^s}$, $t_{prev}^{e^o}$ represent previous time points in history before $t',t''$. $\mathbf{h^{other}}$ stands for hidden layer for the entities (other than the ones in focus) involved in events at $t'$ and $t''$. $r^{e^s}_{prev}=r_2$ and $r^{e^o}_{prev}=r_1$. All other notations mean exactly as defined in text. We only label nodes, edges and embeddings directly relevant to event at time $t$ for clarity.}
\label{fig:evolve_graph}
\vspace{-2mm}
\end{figure*}

\begin{figure*}[ht]
\small
\centering
\begin{tabular}{c|c}
\includegraphics[width = 0.5\textwidth]{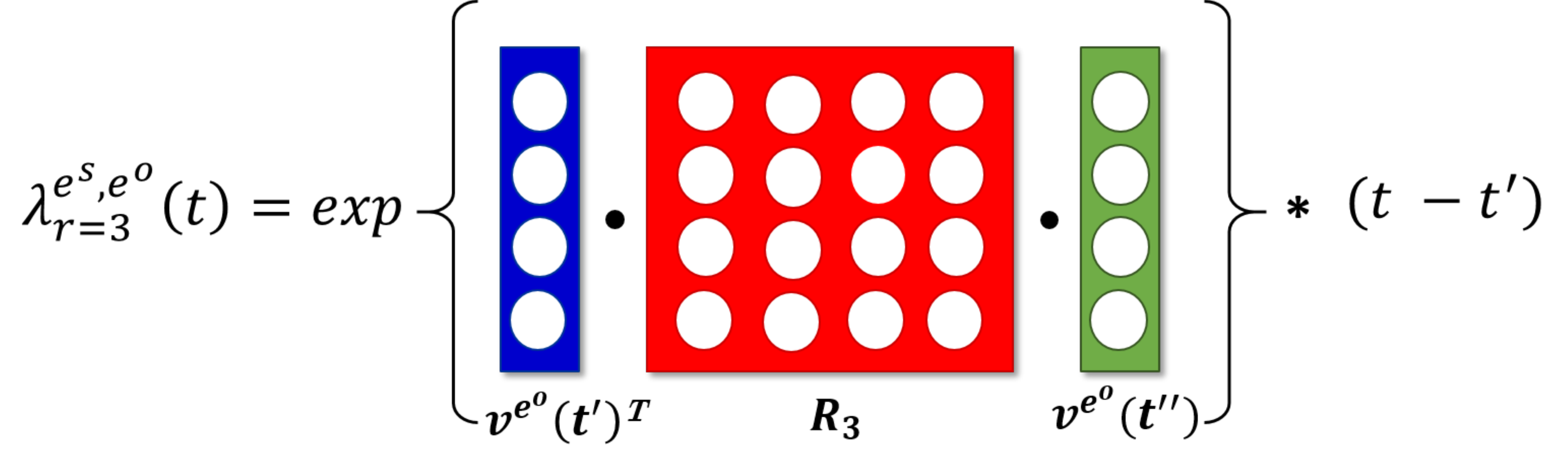}
& \includegraphics[width = 0.5\textwidth]{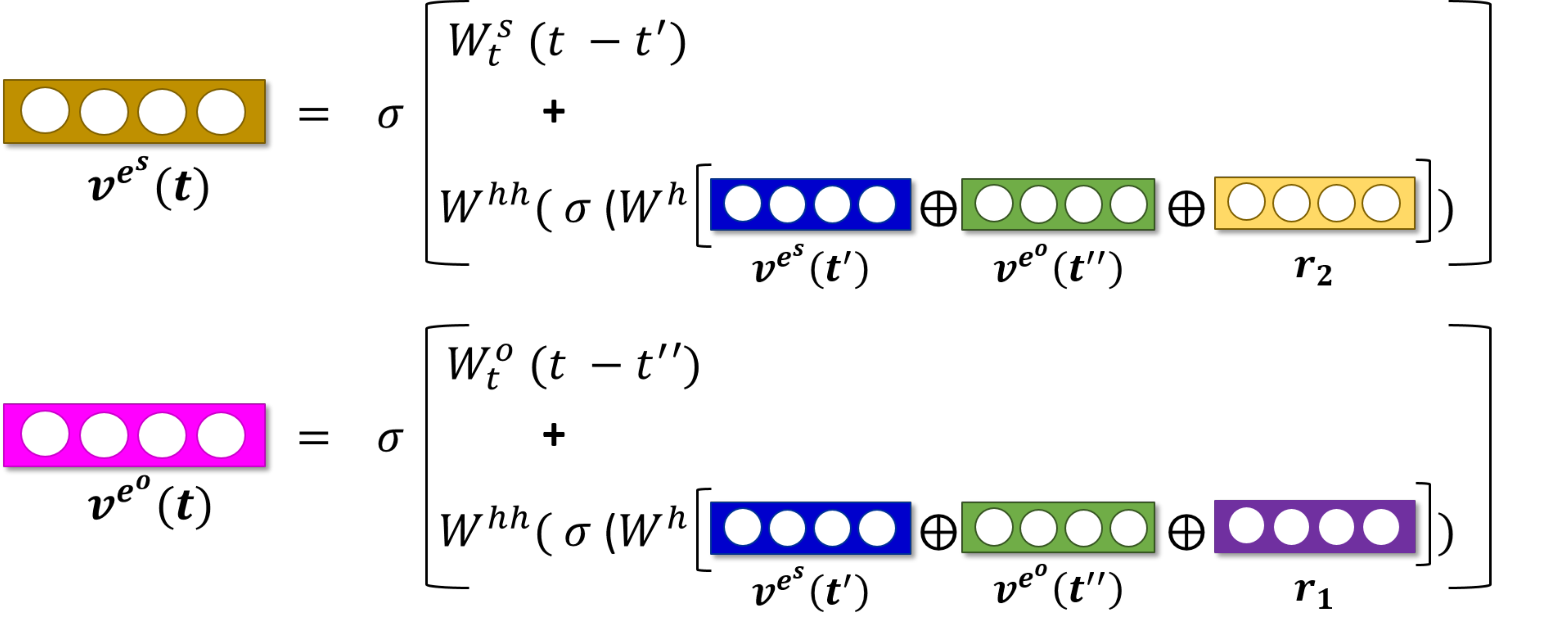}\\
(a) Intensity Computation at time $t$ & (c) Entity Embedding update after event observed at time $t$
\end{tabular}
\vspace{-2mm}
\caption{One step visualization of Know-Evolve computations done in Figure~\ref{fig:evolve_graph} after observing an event at time $t$. (Best viewed in color)}
\label{fig:model}
\vspace{-2mm}
\end{figure*}

We represent latent feature embedding of an entity $e$ at time $t$ with a low-dimensional vector $\mathbf{v^e}(t)$. We add superscript $\mathbf{s}$ and $\mathbf{o}$ as shown in Eq.~(\ref{eq:sc}) to indicate if the embedding corresponds to entity in subject or object position respectively. We also use relationship-specific low-dimensional representation for each relation type.
 
The latent representations of entities change over time as entities forge relationships with each other. We design novel deep recurrent neural network based update functions to capture mutually evolving and nonlinear dynamics of entities in their vector space representations. We consider an event $m = (e^s, r, e^o, t)_m \in \mathcal{D}$ occurring at time $t$. Also, consider that event $m$ is entity $e^s$'s $p$-th event while it is entity $e^o$'s $q$-th event. As entities participate in events in a heterogeneous pattern, it is less likely that $p=q$ although not impossible. Having observed this event, we update the embeddings of two involved entities as follows: 

{\bf Subject Embedding:}  
\begin{equation}
\label{eq:sub}
\begin{split}
\mathbf{v^{e^s}}(t_p) &= \sigma(\mathbf{W^{s}_t}(t_p - t_{p-1}) + \mathbf{W^{hh}} \cdot \mathbf{h^{e^s}}(t_{p}-))\\
\mathbf{h^{e^s}}(t_{p}-) &= \sigma(\mathbf{W^h} \cdot [\mathbf{v^{e^s}}(t_{p-1})\oplus\mathbf{v^{e^o}}(t_p-)\oplus\mathbf{r^{e^s}_{p-1}}])
\end{split}
\vspace{-0.5cm}
\end{equation}
{\bf Object Embedding:}  
\begin{equation}
\label{eq:obj}
\begin{split}
\mathbf{v^{e^o}}(t_q) &= \sigma(\mathbf{W^{o}_t}(t_q - t_{q-1}) + \mathbf{W^{hh}} \cdot \mathbf{h^{e^o}}(t_{q}-))\\
\mathbf{h^{e^o}}(t_{q}-) &= \sigma(\mathbf{W^h} \cdot [\mathbf{v^{e^o}}(t_{q-1})\oplus\mathbf{v^{e^s}}(t_q-)\oplus\mathbf{r^{e^o}_{q-1}}])
\end{split}
\end{equation}
where, $\mathbf{v^{e^s}}$, $\mathbf{v^{e^o}}$ $\in \RR^{d}$. $t_p = t_q = t_m$ is the time of observed event. For subject embedding update in Eq.~(\ref{eq:sub}), $t_{p-1}$ is the time point of the previous event in which entity $e^s$ was involved. $t_p-$ is the timepoint just before time $t_p$. Hence, $\mathbf{v^{e^s}}(t_{p-1})$
represents latest embedding for entity $e^s$ that was updated after $(p-1)$-th event for that entity. $\mathbf{v^{e^o}}(t_p-)$ represents latest embedding for entity $e^o$ that was updated any time just before $t_p = t_m$. This accounts for the fact that entity $e^o$ may have been involved in some other event during the interval between current ($p$) and previous ($p-1$) event of entity $e^s$. $\mathbf{r^{e^s}_{p-1}} \in \RR^{c}$ represent relationship embedding that corresponds to relationship type of the $(p-1)$-th event of entity $e^s$. Note that the relationship vectors are static and do not evolve over time. $\mathbf{h^{e^s}}(t_{p}-) \in \RR^{d}$  is the hidden layer. The semantics of notations apply similarly to object embedding update in Eq.~(\ref{eq:obj}).

$\mathbf{W_t^{s}}, \mathbf{W_t^{o}} \in \RR^{d \times 1}$, $\mathbf{W^{hh}} \in \RR^{d \times l}$ and $\mathbf{W^h} \in \RR^{l \times (2d + c)}$ are weight parameters in network learned during training. $\mathbf{W_t^{s}}, \mathbf{W_t^{o}}$ captures variation in temporal drift for subject and object respectively. $\mathbf{W^{hh}} $ is shared parameter that captures recurrent participation effect for each entity. $\mathbf{W^h}$ is a shared projection matrix applied to consider the compatibility of entities in their previous relationships. $\oplus$ represent simple concatenation operator. $\sigma(\cdot)$ denotes nonlinear activation function ($tanh$ in our case). Our formulations use simple RNN units but it can be replaced with more expressive units like LSTM or GRU in straightforward manner. In our experiments, we choose $d = l$ and $d \neq c$ but they can be chosen differently.
Below we explain the rationales of our deep recurrent architecture that captures nonlinear evolutionary dynamics of entities over time.

{\bf Reasoning Based on Structural Dependency:} The hidden layer ($\mathbf{h^{e^s}}$) reasons for an event by capturing the compatibility of most recent subject embedding with most recent object embedding in previous relationship of subject entity. This accounts for the behavior that within a short period of time, entities tend to form relationships with other entities that have similar recent actions and goals. This layer thereby uses historical information of the two nodes involved in current event and the edges they both created before this event. This holds symmetrically for hidden layer ($\mathbf{h^{e^o}}$).

{\bf Reasoning based on Temporal Dependency:}
The recurrent layer uses hidden layer information to model the intertwined evolution of entity embeddings over time. Specifically this layer has two main components:
\begin{itemize}[leftmargin=*,nosep]
\item {\bf Drift over time:} The first term captures the temporal difference between consecutive events on respective dimension of each entity. This captures the external influences that entities may have experienced between events and allows to smoothly drift their features over time. This term will not contribute anything in case when multiple events happen for an entity at same time point (e.g. within a day in our dataset). While $t_p - t_{p-1}$ may exhibit high variation, the corresponding weight parameter will capture these variations and along with the second recurrent term, it will prevent $\mathbf{v^{e^s}}(t_p)$ to collapse.

\vspace{0.2cm}
\item {\bf Relation-specific Mutual Evolution:} The latent features of both subject and object entities influence each other. In multi-relational setting, this is further affected by the relationship they form. Recurrent update to entity embedding with the information from the hidden layer allows to capture the intricate non-linear and evolutionary dynamics of an entity with respect to itself and the other entity in a specific relationship space. 

\end{itemize}
\vspace{-0.2cm}
\subsection{Understanding Unified View of Know-Evolve}

Figure~\ref{fig:evolve_graph} and Figure~\ref{fig:model} shows the architecture of knowledge evolution framework and one step of our model.

The updates to the entity representations in Eq.~(\ref{eq:sub}) and~(\ref{eq:obj}) are driven by the events involving those entities which makes the embeddings piecewise constant i.e. an  entity embedding remains unchanged in the duration between two events involving that entity and updates only when an event happens on its dimension. This is justifiable as an entity's features may update only when it forges a relationship with other entity within the graph. Note that the first term in Eq.~(\ref{eq:sub}) and~(\ref{eq:obj}) already accounts for any external influences. 

Having observed an event at time $t$, Know-Evolve considers it as an incoming fact that brings new knowledge about the entities involved in that event. It computes the intensity of that event in Eq. (\ref{eq:intent}) which is based on relational compatibility score in Eq. (\ref{eq:sc}) between most recent latent embeddings of involved entities. As these embeddings are piecewise constant, we use time interval term ($t - \bar{t}$) in Eq. (\ref{eq:intent}) to make the overall intensity piecewise linear which is standard mathematical choice for efficient computation in point process framework. This formulation naturally leads to Rayleigh distribution which models time interval between current event and most recent event on either entities' dimension. Rayleigh distribution has an added benefit of having a simple analytic form of likelihood which can be further used to find entity for which the likelihood reaches maximum value and thereby make precise entity predictions. 

\vspace{-0.3cm}
\section{Efficient Training Procedure}

The complete parameter space for the above model is: \\$\mathbf{\Omega} = \{\{\mathbf{V^e}\}_{e=1:n_e}, \{\mathbf{R_r}\}_{r=1:n_r}, \mathbf{W_e},\mathbf{W^{s}_t},\mathbf{W^{o}_t},\mathbf{W^h},$\\$\mathbf{W^{hh}},\mathbf{W_r}\}.$ 
Although Know-Evolve gains expressive power from deep architecture, 
Table 4 (Appendix~\ref{sec:param_comp}) shows that the memory footprint of our model is comparable to simpler relational models.
The intensity function in (\ref{eq:intent}) allows to use maximum likelihood estimation over all the facts as our objective function. Concretely, given a collection of facts recorded in a temporal window $[0,T)$, we learn the model by minimizing the joint negative log likelihood of intensity function ~\citep{DalVer2007} written as:
\vspace{-0.3cm}
\begin{equation}
\label{eq:data_loglikelihood}
\begin{split}
\mathcal{L} &= \underbrace{- \sum_{p=1}^N \log\rbr{\lambda^{e^s,e^o}_r(t_p|\bar{t_p})}}_{\text{happened events}}\\ &+ \underbrace{\sum_{r=1}^{n_r}\sum_{e^s=1}^{n_e}\sum_{e^o=1}^{n_e}\int_0^T \lambda^{e^s,e^o}_{r}(\tau|\bar{\tau})\, d \tau}_{\text{survival term}}
\end{split}
\end{equation}
The first term maximizes the probability of specific type of event between two entities; the second term  penalizes non-presence of all possible types of events between all possible entity pairs in a given observation window. We  use  Back Propagation Through Time (BPTT) algorithm to train our model. 
Previous techniques ~\citep{DuDaiTriUpaGomSon16,HidKarBalTik16} that use BPTT algorithm decompose data into independent sequences and train on mini-batches of those sequences. But there exists intricate relational and temporal dependencies between data points in our setting which limits our ability to efficiently train by decomposing events into independent sequences. To address this challenge, we design an efficient Global BPTT algorithm (Algorithm~\ref{alg:alg1},  Appendix~\ref{sec:alg2}) that   creates mini-batches of events over global timeline in sliding window fashion and allows to capture dependencies across batches while retaining efficiency.

 
{\bf Intractable Survival Term.} 
To compute the second survival term in (\ref{eq:data_loglikelihood}), since our intensity function is modulated by relation-specific parameter, for each relationship we need to compute survival probability over all pairs of entities. Next, given a relation $r$ and entity pair $(e^s,e^o)$, we denote $P_{(e^s, e^o)}$ as total number of events of type $r$ involving either $e^s$ or $e^o$  in window [$T_0$, $T$). As our intensity function is piecewise-linear, we can decompose the integration term $-\int_{T_0}^T \lambda^{e^s,e^o}_r (\tau | \bar{\tau})d\tau$  into multiple time intervals where intensity is constant:
\vspace{-0.2cm}
\begin{align}
\label{eq:surv}
& \int_{T_0}^T \lambda^{e^s,e^o}_r (\tau | \bar{\tau})d\tau \nonumber \\&=  \sum_{p=1}^{P_{(e^s, e^o)} -1} \int_{t_p}^{t_{p + 1}} \lambda^{e^s,e^o}_r (\tau | \bar{\tau})d\tau \nonumber
\\&= \sum_{p=1}^{P_{(e^s, e^o)} -1}(t^2_{p+1} - t^2_{p}) \cdot exp(\mathbf{v^{e^s}}(t_p)^T\cdot\mathbf{R_r}\cdot\mathbf{v^{e^o}}(t_p))	
\end{align}
\vspace{-0.5cm}

The integral calculations in (\ref{eq:surv}) for all possible triplets requires $\mathcal{O}(n^2r)$ computations ($n$ is number of entities and $r$ is the number of relations). This is computationally intractable and also unnecessary. Knowledge tensors are inherently sparse and hence it is plausible to approximate the survival loss in a stochastic setting.
We take inspiration from techniques like noise contrastive \citep{GutHyv12} estimation and adopt a random sampling strategy to compute survival loss: Given a mini-batch of events, for each relation in the mini-batch, we compute dyadic survival term across all entities in that batch. Algorithm~\ref{alg:alg2} presents the survival loss computation procedure. While this procedure may randomly avoid penalizing some dimensions in a relationship, it still includes all dimensions that had events on them. The computational complexity for this procedure will be $\mathcal{O}(2n'r'm)$ where $m$ is size of mini-batch and $n'$ and $r'$ represent number of entities and relations in the mini-batch.
\vspace{-0.5cm}
\setlength{\textfloatsep}{0.2cm}
\begin{algorithm}[t]
   \caption{Survival Loss Computation in mini-batch}
   \label{alg:alg2}
\begin{algorithmic}
   \STATE {\bfseries Input:} Mini\-batch $\mathcal{E}$, size $s$, Batch Entity List $bl$
   \STATE $loss=0.0$
   \FOR{$p=0$ {\bfseries to} $s-1$}
   \STATE subj\_feat = $\mathcal{E}_p \rightarrow \mathbf{v^{e^s}}(t-)$
   \STATE obj\_feat = $\mathcal{E}_p \rightarrow \mathbf{v^{e^o}}(t-)$
   \STATE rel\_weight = $\mathcal{E}_p \rightarrow \mathbf{R_r}$
   \STATE t\_end = $\mathcal{E}_p \rightarrow t$
   \STATE $subj\_surv = 0$, $obj\_surv = 0$, $total\_surv = 0$
   \FOR{$i=0$ {\bfseries to} $bl.size$}
   \STATE obj\_other = $bl[i]$
   \IF{obj\_other $== \mathcal{E}_p \rightarrow e^s$} 
   \STATE continue
   \ENDIF
   \STATE $\bar{t} = max(t^{e^s}-,t^{e^o}-)$
   \STATE $subj\_surv \mathrel{+}= (t\_end^2 - \bar{t}^2) \cdot exp(subj\_feat^T\cdot rel\_weight \cdot obj\_other\_feat)$
   \ENDFOR
   \FOR{$j=0$ {\bfseries to} $bl.size$}
   \STATE subj\_other = $bl[i]$
   \IF{subj\_other $== \mathcal{E}_p \rightarrow e^o$} 
   \STATE continue
   \ENDIF
   \STATE $\bar{t} = max(t^{e^s}-,t^{e^o}-)$
   \STATE $obj\_surv \mathrel{+}= (t\_end^2 - \bar{t}^2) \cdot exp(subj\_other\_feat^T\cdot rel\_weight \cdot obj\_feat)$
   \ENDFOR
   \STATE $loss \mathrel{+}= subj\_surv + obj\_surv$
   \ENDFOR
\end{algorithmic}
\end{algorithm}

\vspace{-0.2cm}
\section{Experiments}

\begin{figure*} [ht!]
\small
\centering
\begin{tabular}{cccc}
\includegraphics[width = 0.20\textwidth]{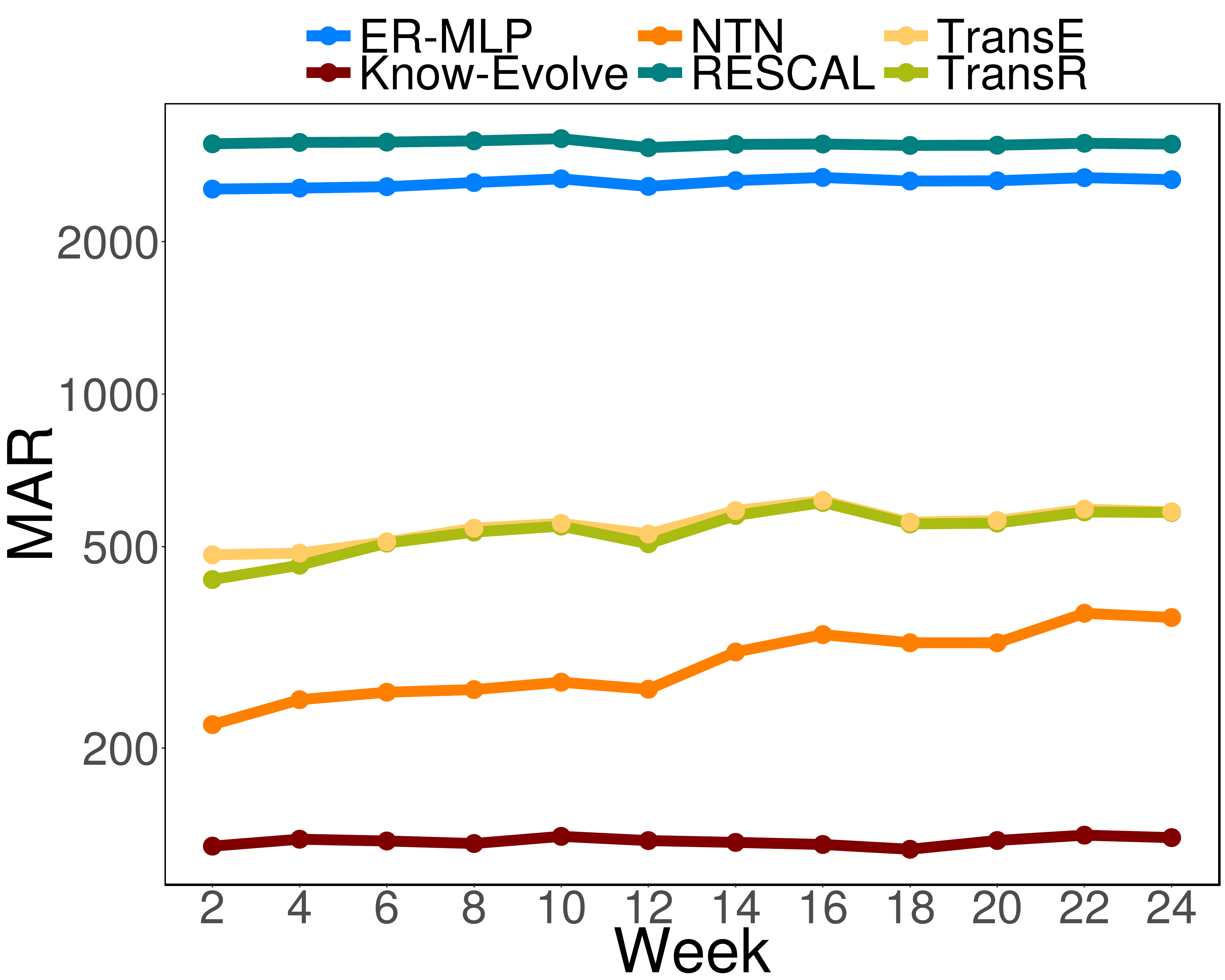}
& \includegraphics[width = 0.20\textwidth]{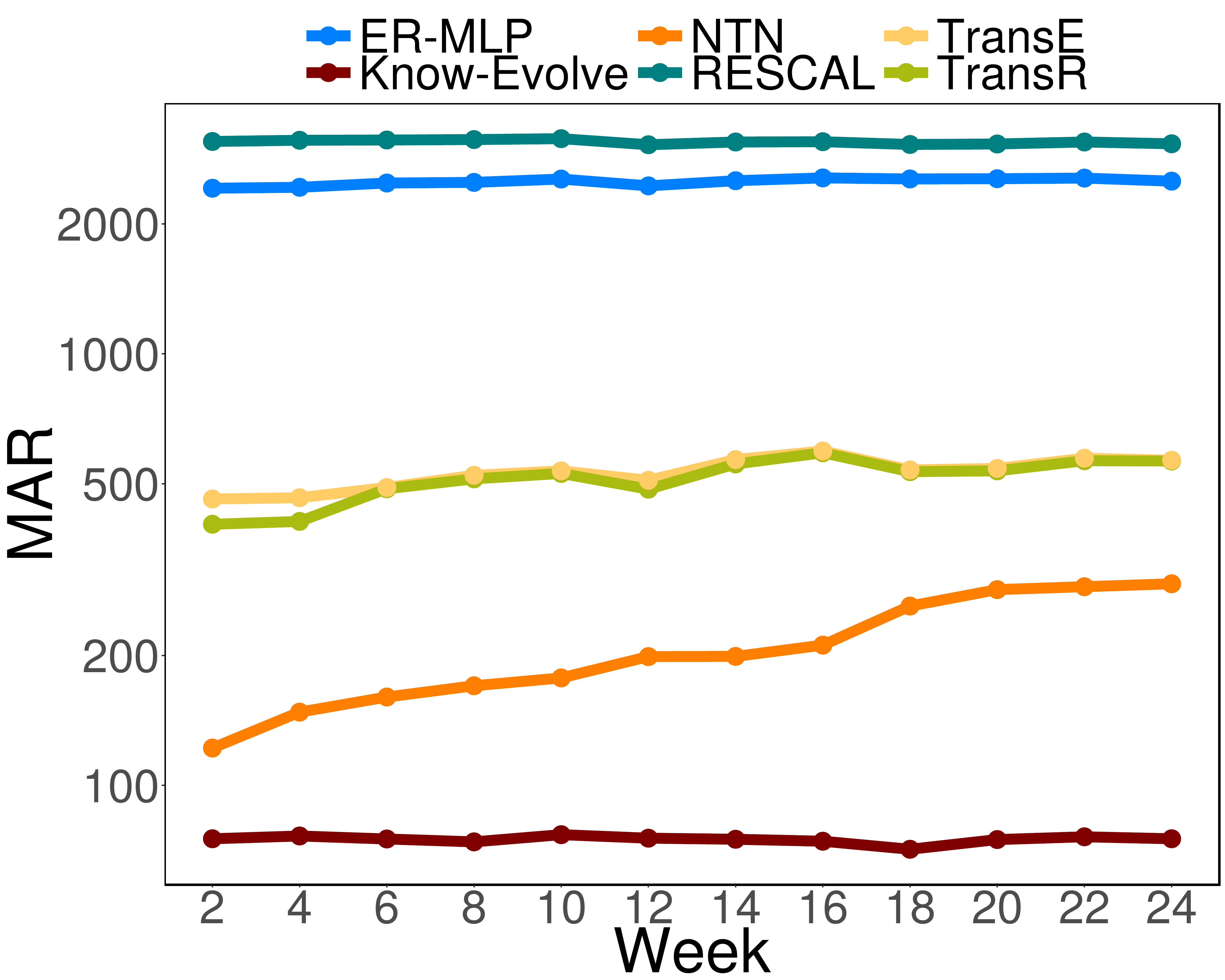}
& \includegraphics[width = 0.20\textwidth]{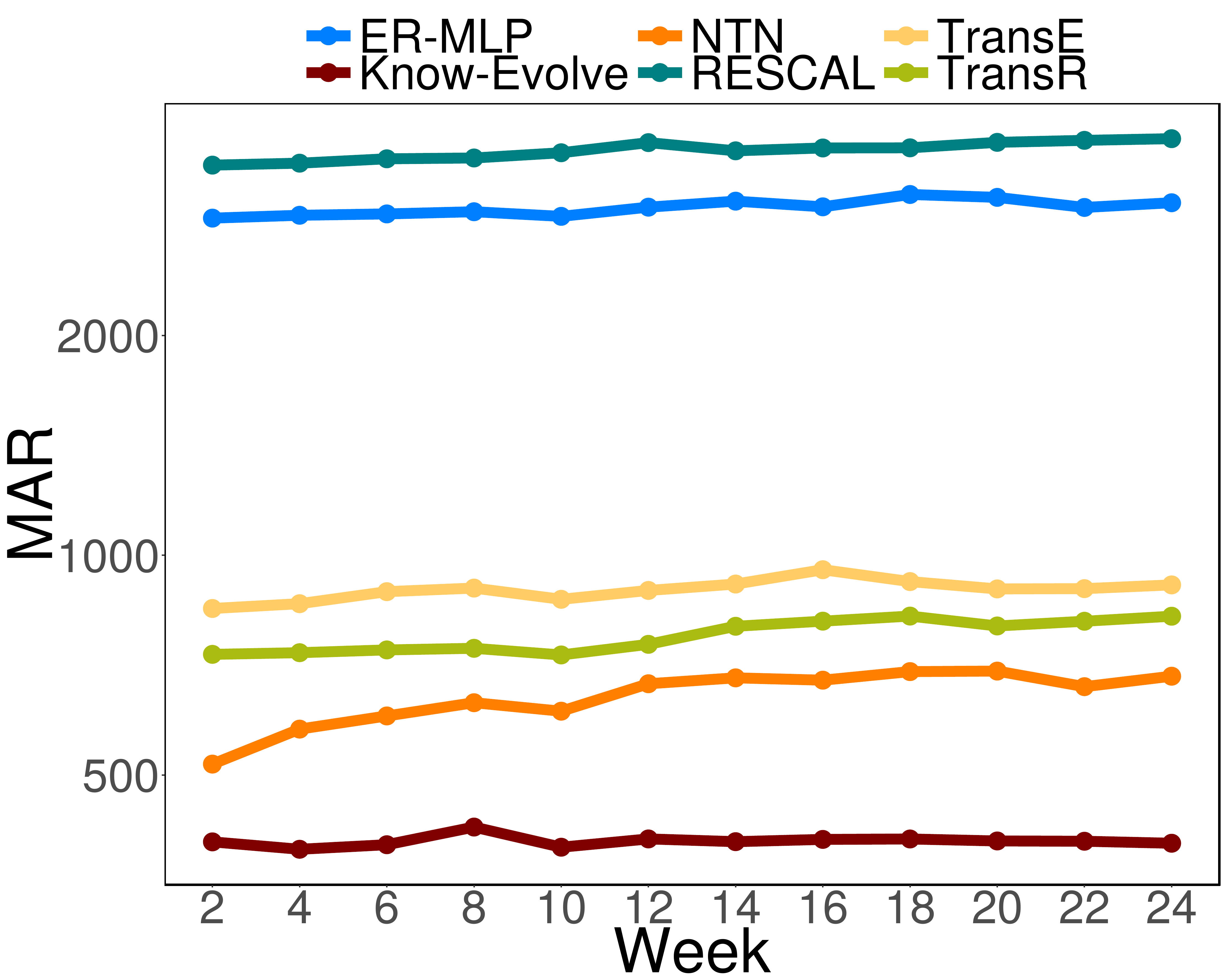}
& \includegraphics[width = 0.20\textwidth]{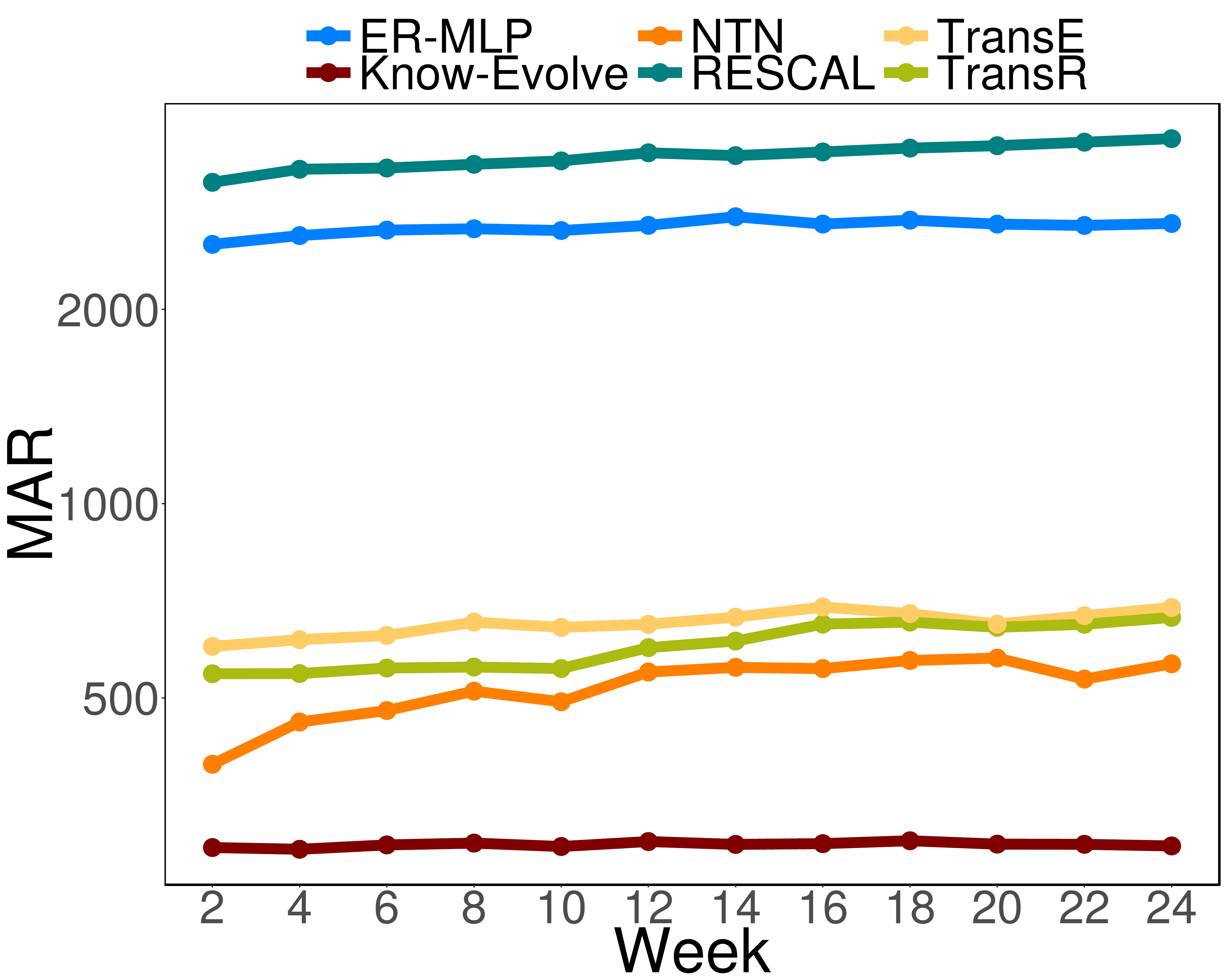}\\
(a) ICEWS-raw & (b) ICEWS-filtered & (c) GDELT-raw & (d) GDELT-filtered
\end{tabular}
\vspace{-2mm}
\caption{Mean Average Rank (MAR) for Entity Prediction on both datasets.}
\label{fig:mar}
\vspace{-0.1cm}
\end{figure*}

\begin{figure*} [ht!]
\small
\centering
\begin{tabular}{cccc}
\includegraphics[width = 0.20\textwidth]{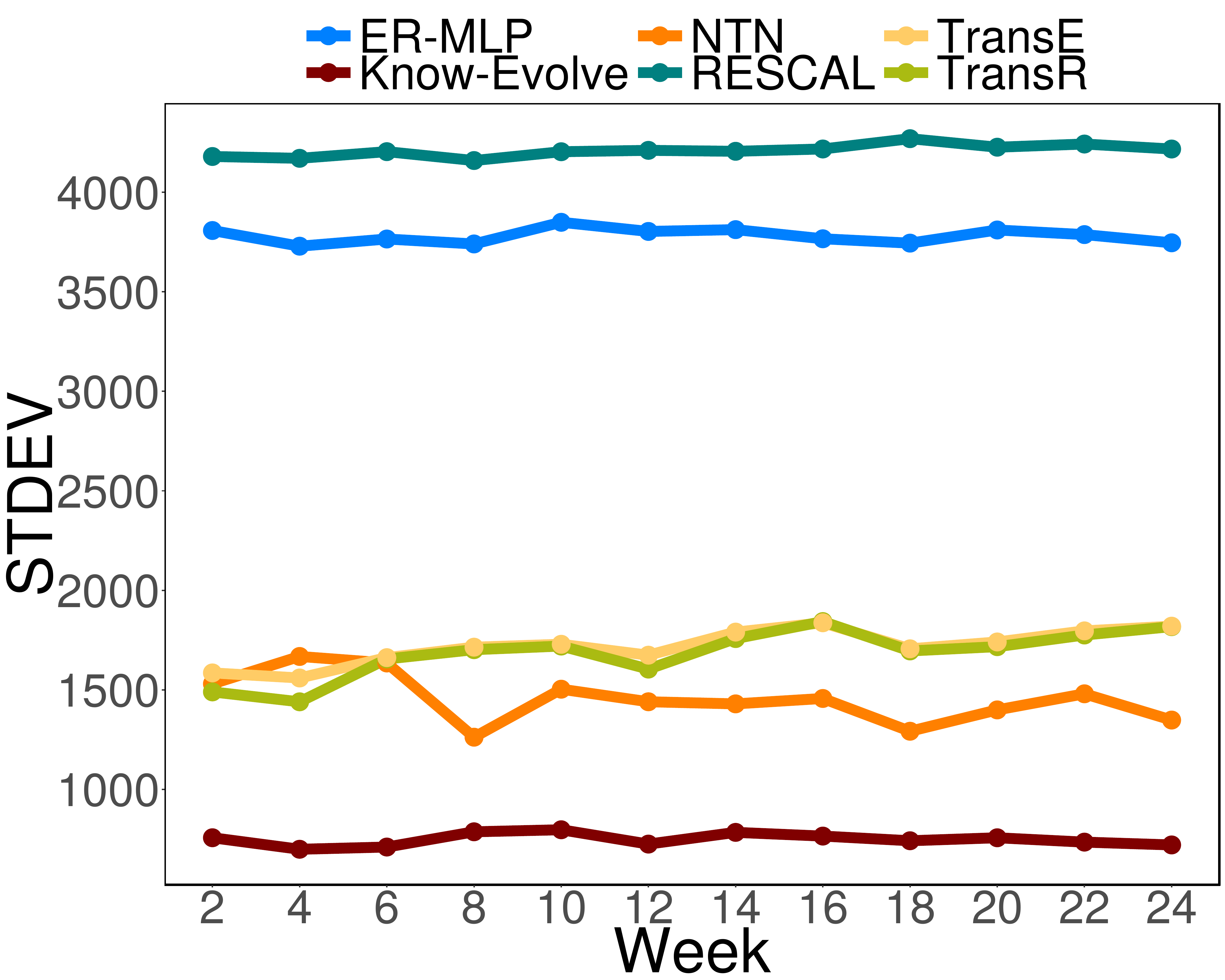}
& \includegraphics[width = 0.20\textwidth]{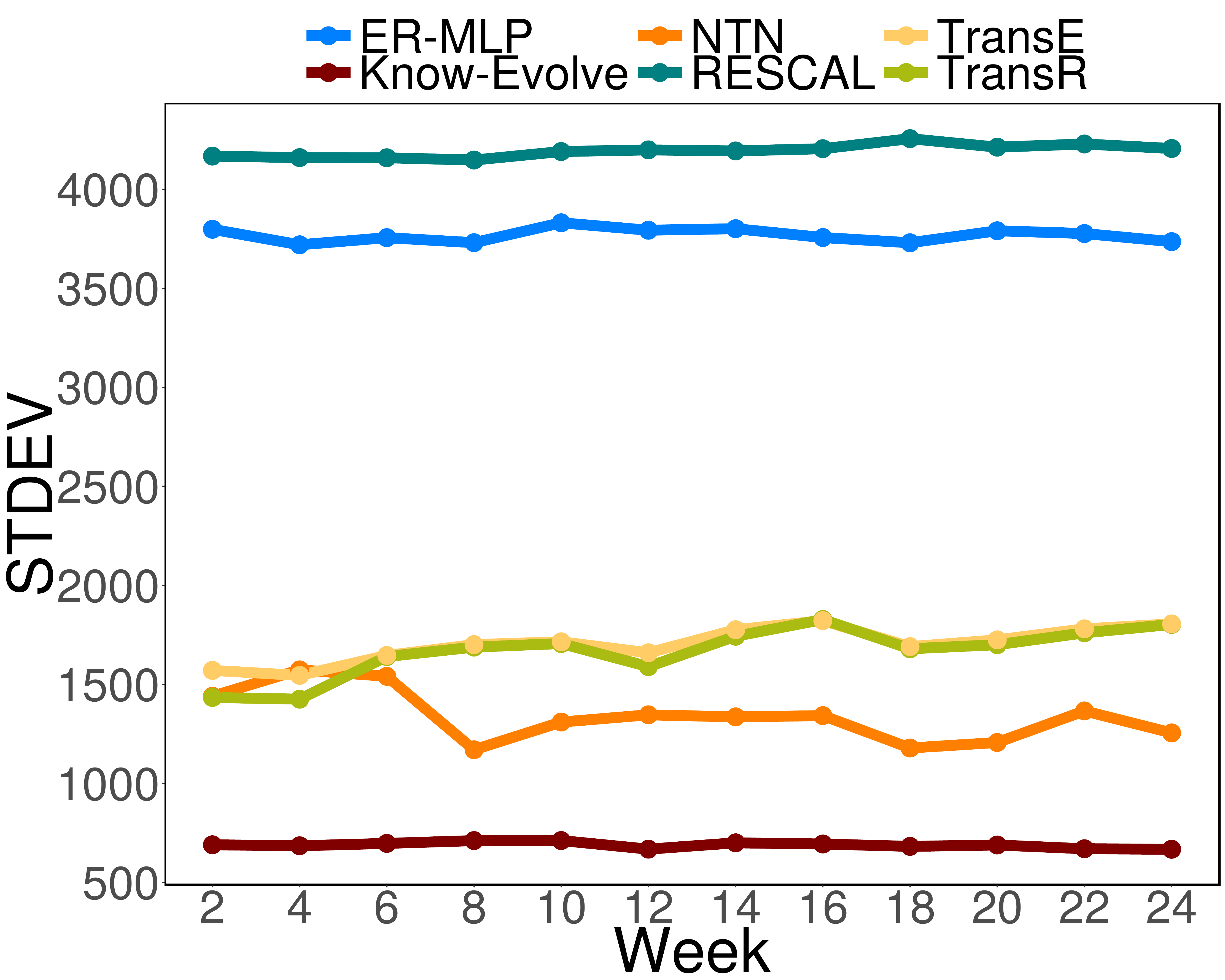}
& \includegraphics[width = 0.20\textwidth]{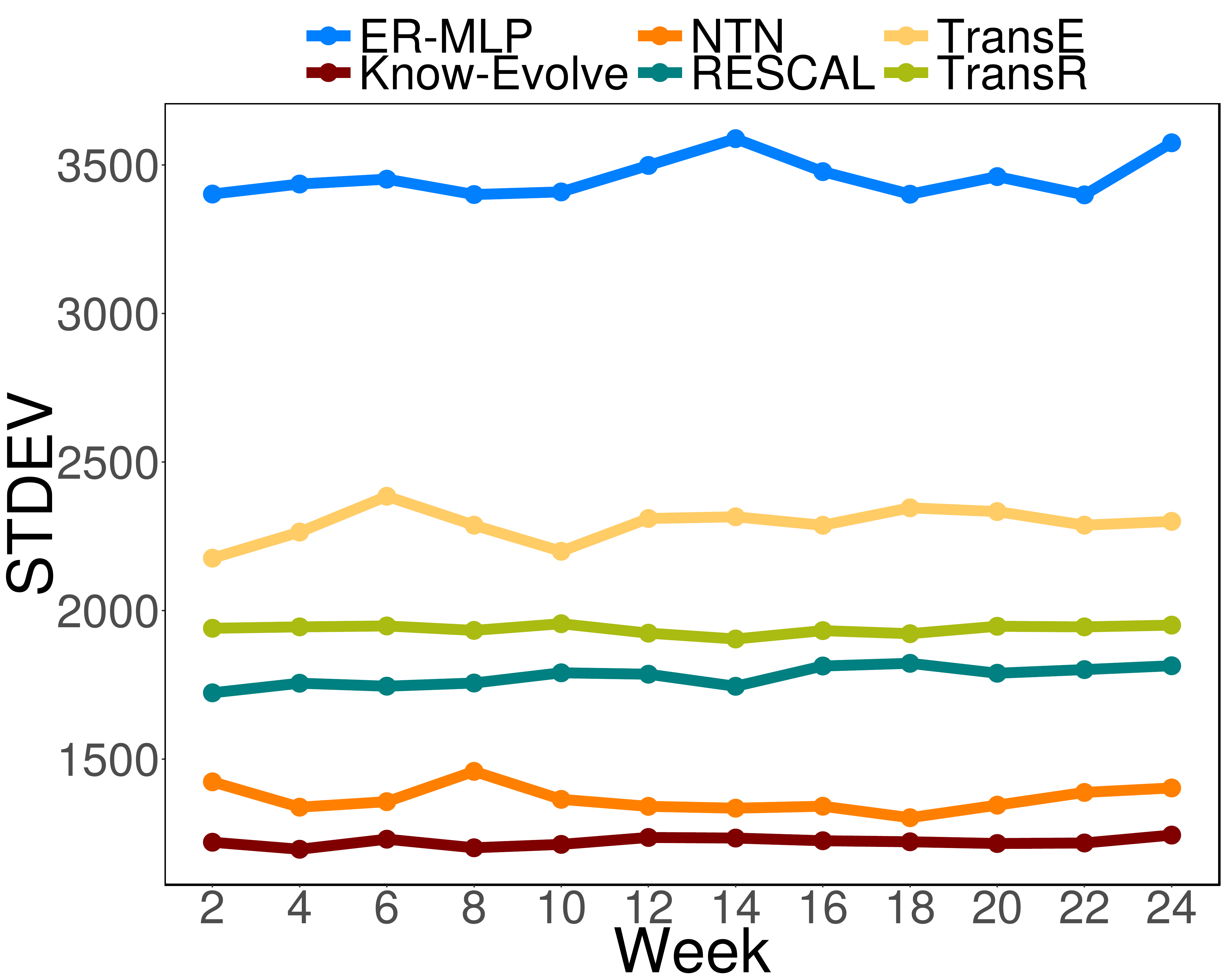}
& \includegraphics[width = 0.20\textwidth]{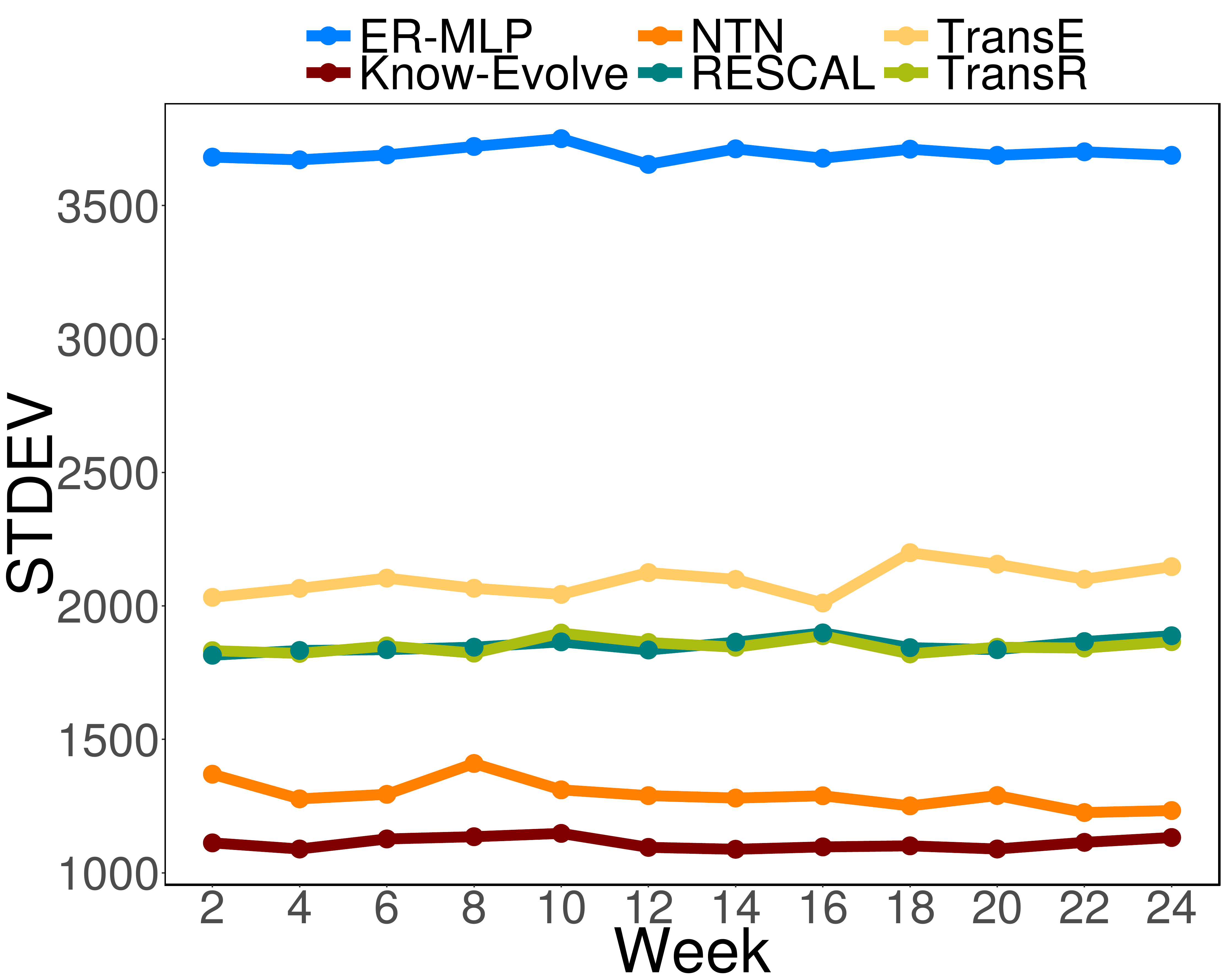}\\
(a) ICEWS-raw & (b) ICEWS-filtered & (c) GDELT-raw & (d) GDELT-filtered
\end{tabular}
\vspace{-2mm}
\caption{Standard Deviation (STD) in MAR for Entity Prediction on both datasets.}
\label{fig:stdev}
\vspace{-0.1cm}
\end{figure*}

\begin{figure*} [ht!]
\small
\centering
\begin{tabular}{cccc}
\includegraphics[width = 0.20\textwidth]{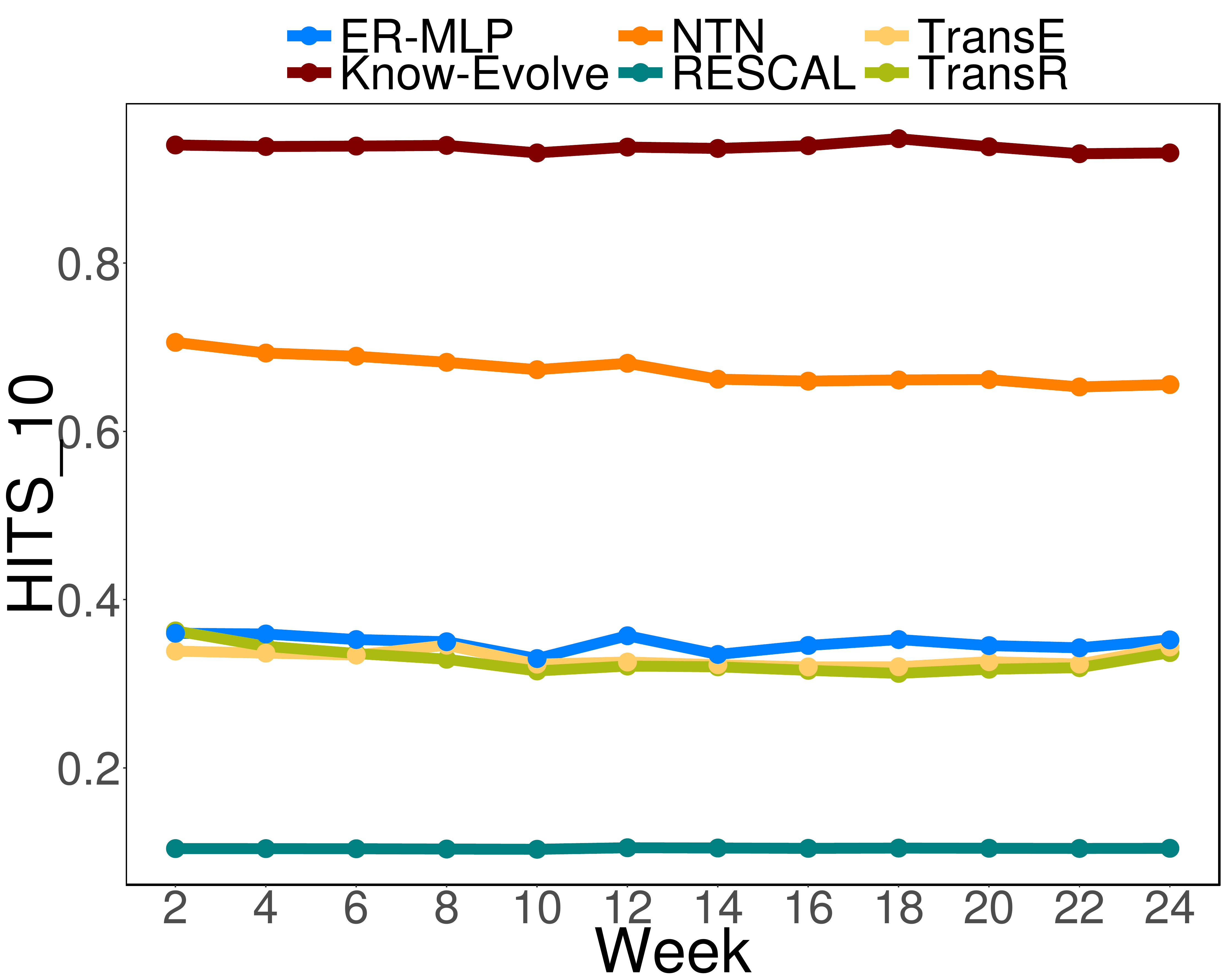}
& \includegraphics[width = 0.20\textwidth]{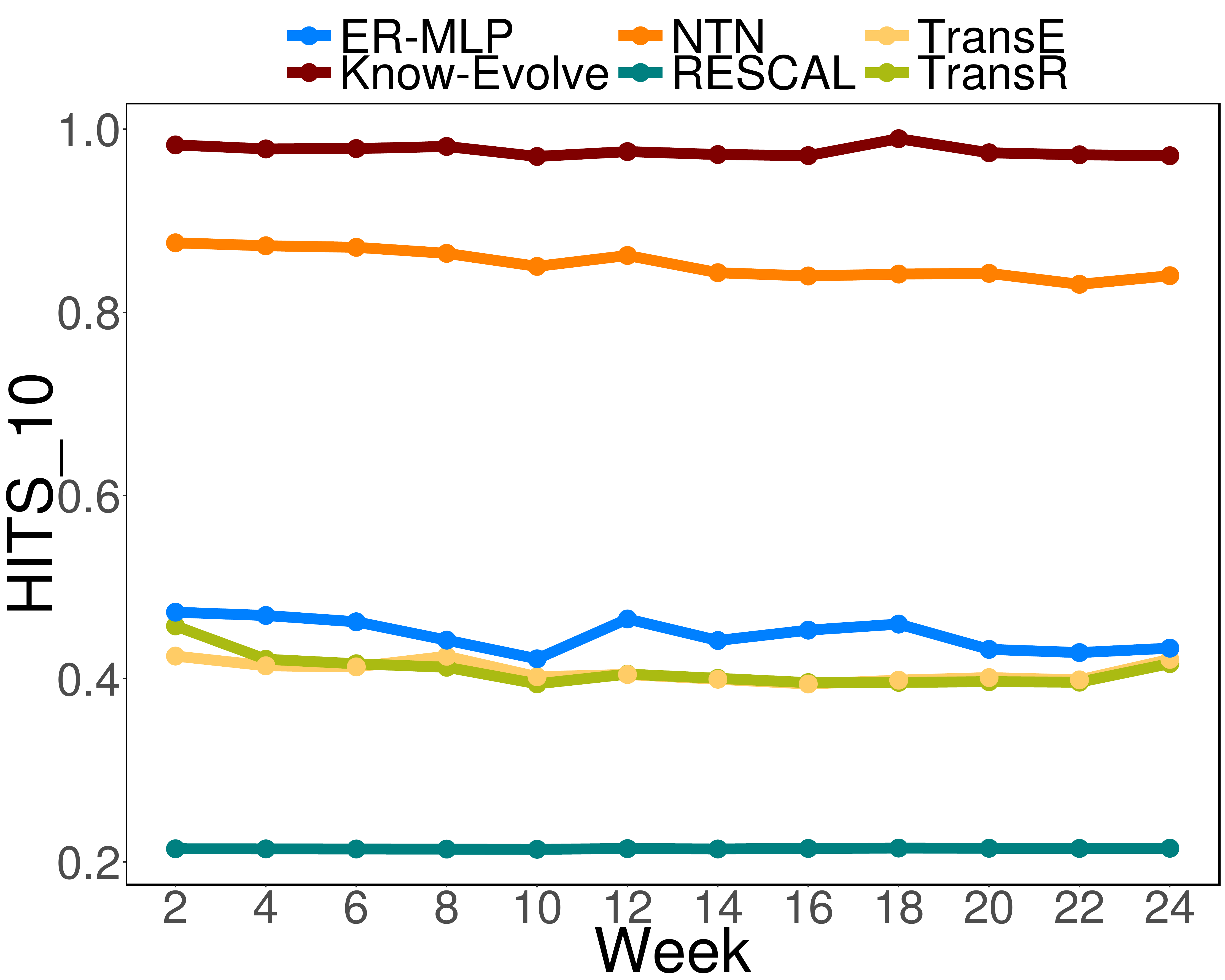}
& \includegraphics[width = 0.20\textwidth]{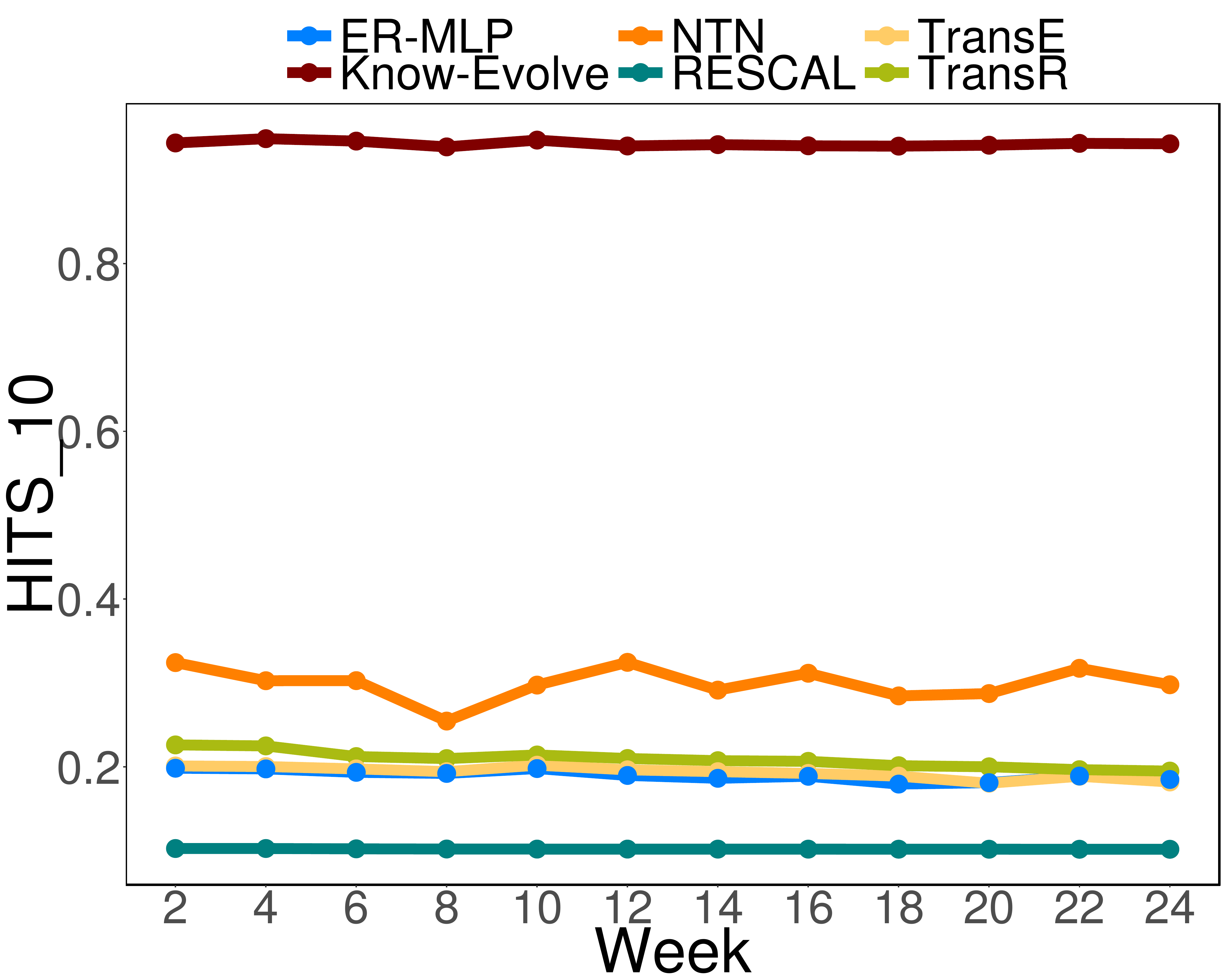}
& \includegraphics[width = 0.20\textwidth]{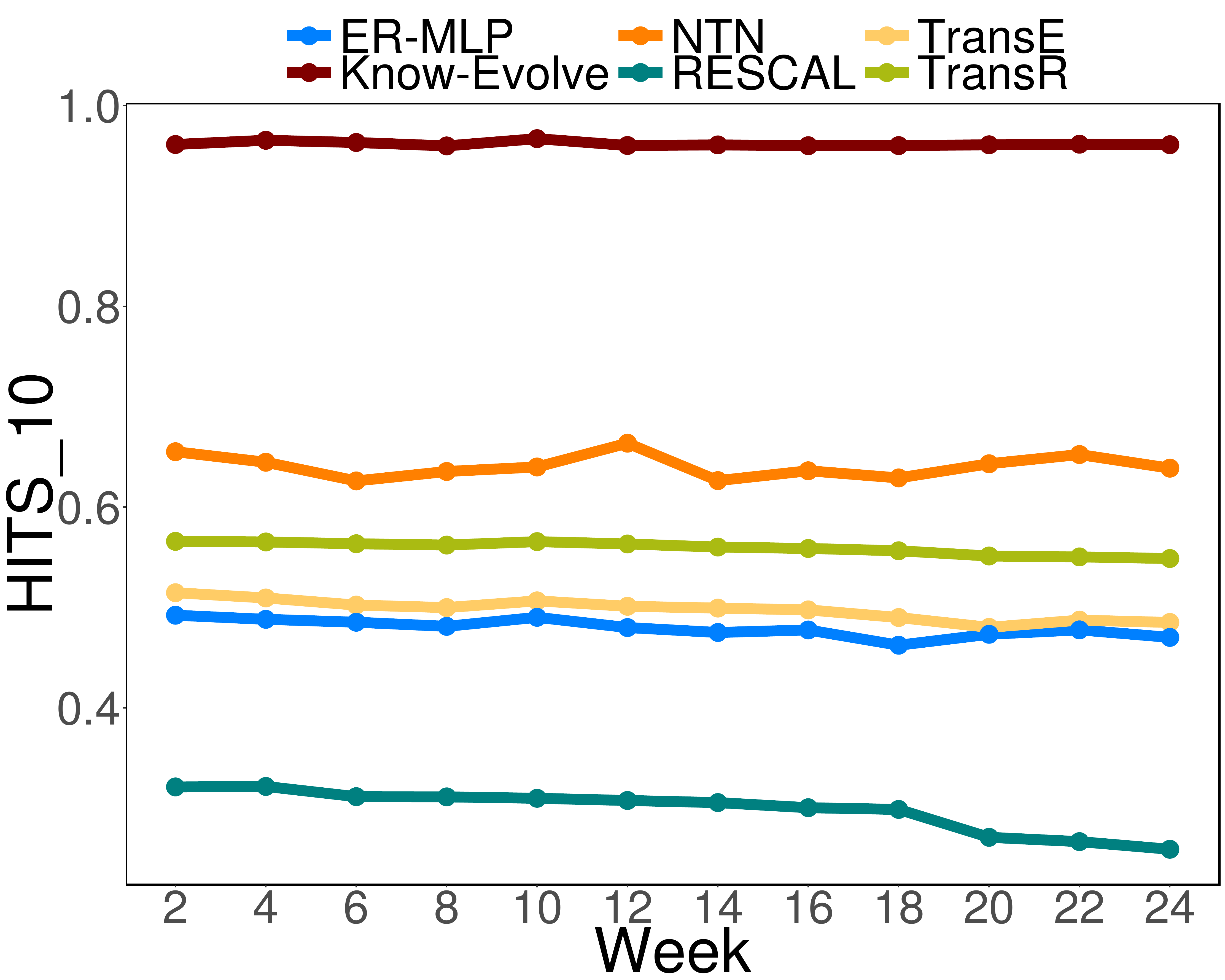}\\
(a) ICEWS-raw & (b) ICEWS-filtered & (c) GDELT-raw & (d) GDELT-filtered
\end{tabular}
\vspace{-2mm}
\caption{HITS@10 for Entity Prediction on both datasets.}
\label{fig:hits10}
\vspace{-0.1cm}
\end{figure*}

\subsection{Temporal Knowledge Graph Data}

We use two datasets: Global Database of Events, Language, and Tone (GDELT) ~\cite{LeeSch13} and Integrated Crisis Early Warning System (ICEWS) ~\cite{BosLauObrSheStaWar17} which has recently gained attention in learning community~\cite{SchZhoMinBleWal16} as useful temporal KGs. GDELT data is collected from April 1, 2015 to Mar 31, 2016 (temporal granularity of 15 mins). ICEWS dataset is collected from Jan 1, 2014 to Dec 31, 2014 (temporal granularity of 24 hrs). Both datasets contain records of events that include two actors, action type and timestamp of event. We use different hierarchy of actions in two datasets - (top level 20 relations for GDELT while last level 260 relations for ICEWS) - to test on variety of knowledge tensor configurations. Note that this does not filter any record from the dataset. We process both datasets to remove any duplicate quadruples, any mono-actor events (\ie, we use only dyadic events), and  self-loops. We report our main results on full versions of each dataset. We create 
smaller version of both datasets for exploration purposes. Table~\ref{tab:data_stat} (Appendix~\ref{sec:stats}) provide statistics about the data and Table~\ref{tab:spar_stat} (Appendix~\ref{sec:stats}) demonstrates the sparsity of knowledge tensor.
%
%
%
\vspace{-0.2cm}
\subsection{Competitors}
We compare the performance of our method with following relational learning methods: 
RESCAL, Neural Tensor Network (NTN), Multiway Neural Network (ER-MLP), TransE and TransR. To the best of our knowledge, there are no existing relational learning approaches that can predict time for a new fact. Hence we devised two baseline methods for evaluating time prediction performance --- \emph{(i) Multi-dimensional Hawkes process (MHP):} We model dyadic entity interactions as multi-dimensional Hawkes process similar to ~\citep{DuWanHeetal15}. Here, an entity pair constitutes a dimension and for each pair we collect sequence of events on its dimension and train and test on that sequence. Relationship is not modeled in this setup. 
\emph{(ii) Recurrent Temporal Point Process (RTPP):} We implement a simplified version of RMTPP ~\citep{DuDaiTriUpaGomSon16} where we do not predict the marker. For training, we concatenate static entity and relationship embeddings and augment the resulting vector with temporal feature. This augmented unit is used as input to global RNN which produces output vector $\mathbf{h_t}$. During test time, for a given triplet, we use this vector $\mathbf{h_t}$ to compute conditional intensity of the event given history which is further used to predict next event time. Appendix~\ref{sec:impl} provides implementation details of our method and competitors.

\vspace{-0.2cm}

\subsection{Evaluation Protocol}
We report experimental results on two tasks: \emph{Link  prediction} and \emph{Time prediction}.

{\bf Link prediction:} Given a test quadruplet $(e^s,r,e^o,t)$, we replace $e^o$ with all the entities in the dataset and compute the conditional density $d^{e^s,e^o}_r = \lambda^{e^s,e^o}_r(t)S^{e^s,e^o}_r(t)$ for the resulting quadruplets including the ground truth. We then sort all the quadruplets in the descending order of this density to rank the correct entity for object position. 
We also conduct testing after applying the filtering techniques described in ~\citep{BorUsuGarWesetal13} - we only rank against the entities that do not generate a true triplet (seen in train) when it replaces ground truth object. 
We report Mean Absolute Rank (MAR), Standard Deviation for MAR and HITS@10 (correct entity in top 10 predictions) for both Raw and Filtered Versions. 

{\bf Time prediction:} Give a test triplet $(e^s,r,e^o)$, we predict the expected value of next time the fact $(e^s,r,e^o)$  can occur. This expectation is defined by: $\mathbb{E}^{e^s, e^o}_r(t) = \sqrt{\frac{\pi}{2\exp\left(g^{e^s,e^o}_r(t)\right)}}$, where $g^{e^s,e^o}_r(t)$ is computed using equation (\ref{eq:sc}). We report Mean Absolute Error (MAE) between the predicted time and true time in hours.

{\bf Sliding Window Evaluation.} As our work concentrates on temporal knowledge graphs, it is more interesting to see the performance of methods over time span of test set as compared to single rank value. This evaluation method can help to realize the effect of modeling temporal and evolutionary knowledge. We therefore partition our test set in 12 different slides and report results in each window. For both datasets, each slide included 2 weeks of time. 

%

\vspace{-0.3cm}
\subsection{Quantitative Analysis}

{\bf Link Prediction Results.} Figure~\ref{fig:mar}, \ref{fig:stdev} and \ref{fig:hits10} demonstrate link prediction performance comparison on both datasets. Know-Evolve significantly and consistently outperforms all competitors in terms of prediction rank without any deterioration over time. Neural Tensor Network's second best performance compared to other baselines demonstrate its rich expressive power but it fails to capture the evolving dynamics of intricate dependencies over time. This is further substantiated by its decreasing performance as we move test window further in time. 

The second row represents deviation error for MAR across samples in a given test window. Our method achieves significantly low deviation error compared to competitors making it most stable. Finally, high performance on HITS@10 metric demonstrates extensive discriminative ability of Know-Evolve. For instance, GDELT has only 20 relations but 32M events where many entities interact with each other in multiple relationships. In this complex setting, other methods depend only on static entity embeddings to perform prediction unlike our method which does effectively infers new knowledge using powerful evolutionary network and provides accurate prediction results. 

%

\begin{figure}[t!]
\small
\centering
\resizebox{0.45\textwidth}{!}{
\begin{tabular}{cc}
\includegraphics[width = 0.22\textwidth]{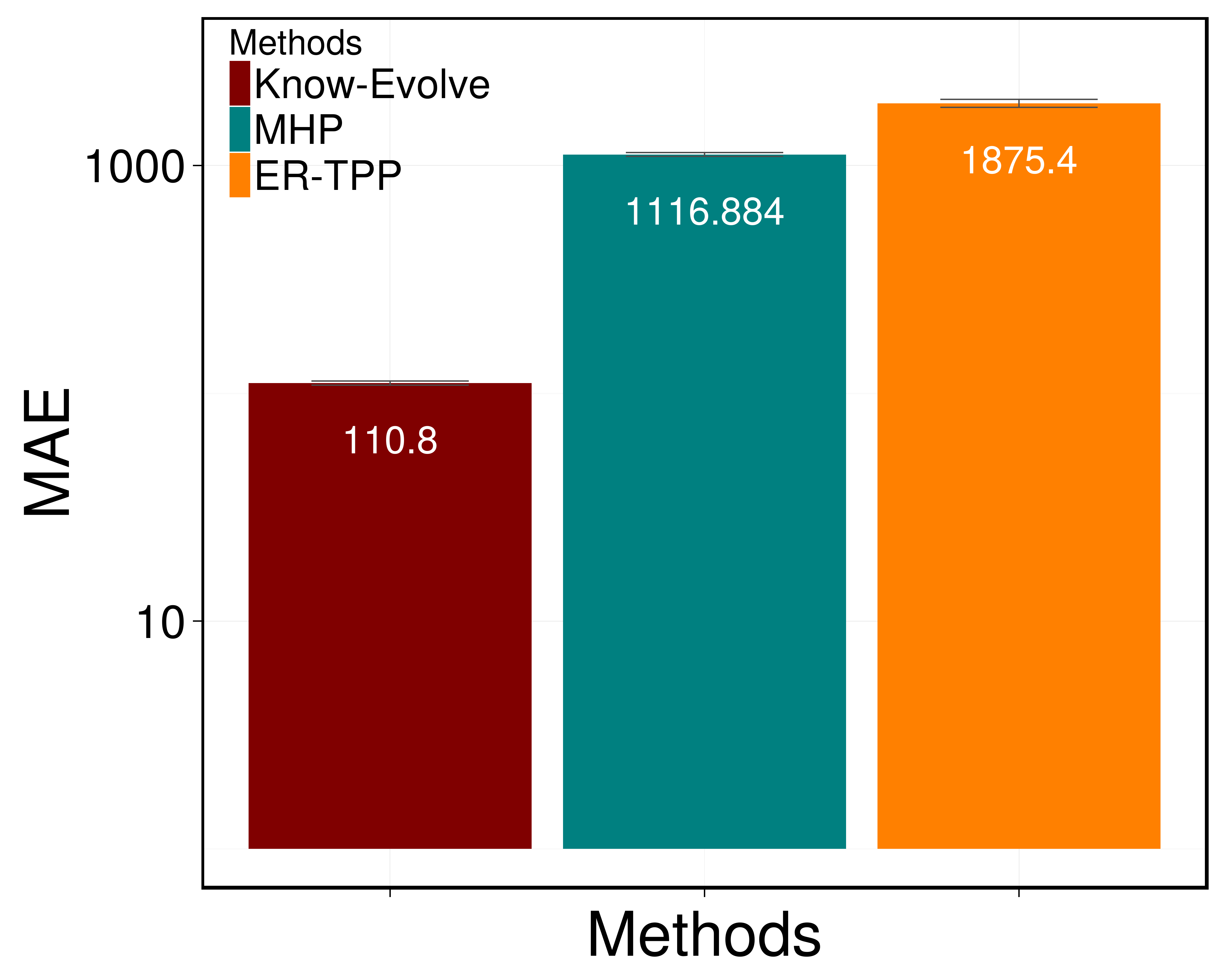}
& \includegraphics[width = 0.22\textwidth]{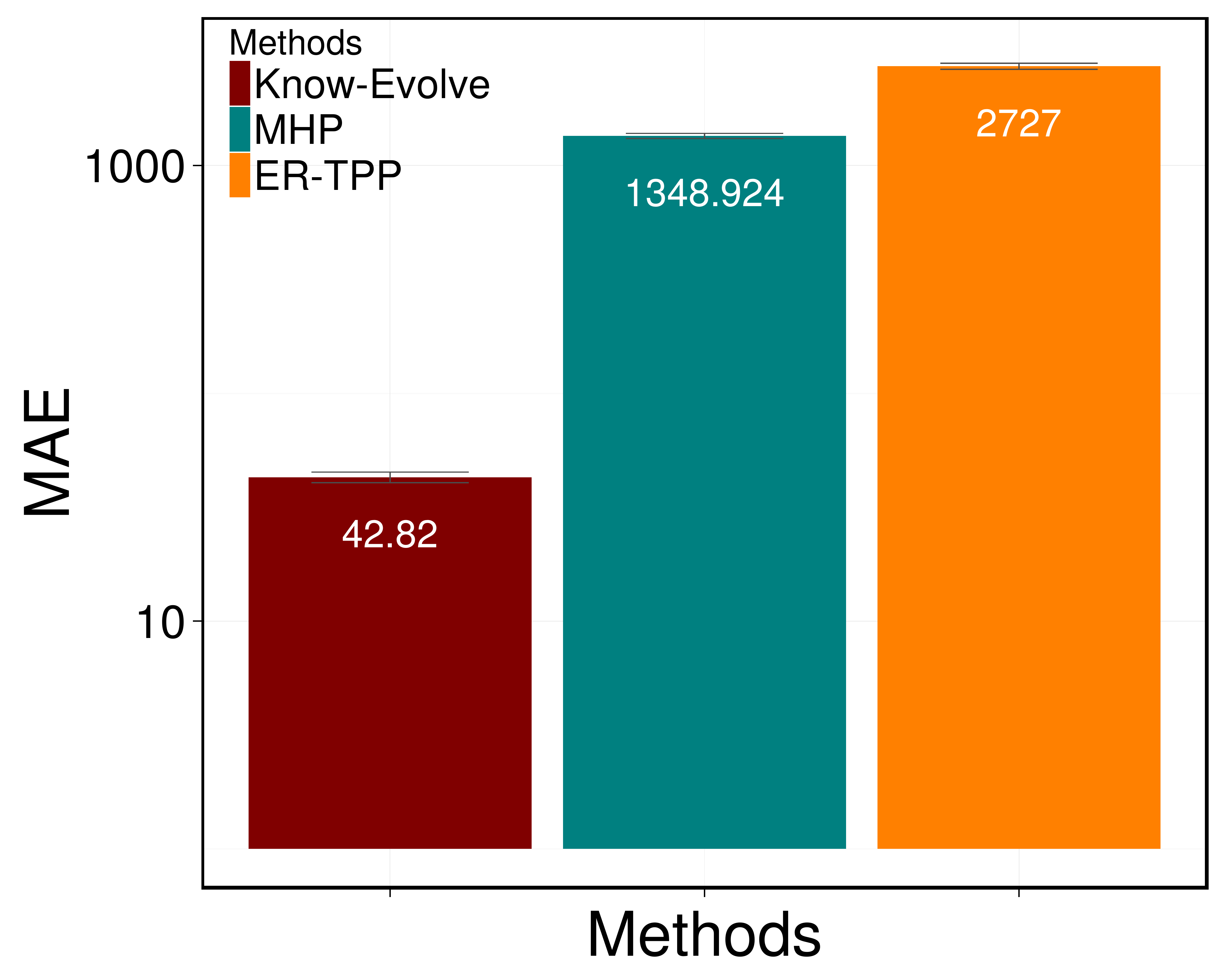}\\
(a) GDELT-500 & (b) ICEWS-500
\end{tabular}
}
\caption{Time prediction performance (Unit is hours).}
\label{fig:time_pred}
\end{figure}

{\bf Time Prediction Results.} Figure~\ref{fig:time_pred} demonstrates that Know-Evolve performs significantly better than other point process based methods for predicting time. MHP uses a specific parametric form of the intensity function which limits its expressiveness. Further, each entity pair interaction is modeled as an independent dimension and does not take into account relational feature which fails to capture the intricate influence of different entities on each other. On the other hand, RTPP uses  relational features as part of input, but it sees all events globally and cannot model the intricate evolutionary dependencies on past events. We observe that our method effectively captures such non-linear relational and temporal dynamics. 

In addition to the superior quantitative performance, we demonstrate the effectiveness of our method by providing extensive exploratory analysis  in Appendix~\ref{sec:Expl}.
\vspace{-0.3cm}

\section{Related Work}
\vspace{-0.1cm}
In this section, we discuss relevant works in relational learning and temporal modeling techniques.
\vspace{-0.2cm}
\subsection{Relational Learning}
Among various relational learning techniques, neural embedding models that focus on learning low-dimensional representations of entities and relations have shown state-of-the-art performance. These methods compute a score for the fact based on different operations on these latent representations. Such models can be mainly categorized into two variants: 

{\bf Compositional Models.} RESCAL~\citep{NicTreKri11} uses a relation specific weight matrix to explain triplets via pairwise interactions of latent features. Neural Tensor Network (NTN)~\citep{SocCheManNg13} is more expressive model as it combines a standard NN layer with a bilinear tensor layer. ~\citep{DonGabHeiHorLaoMurStrSunZha14} employs a concatenation-projection method to project entities and relations to lower dimensional space. Other sophisticated models include Holographic Embeddings (HoLE) ~\citep{NicRosPog16} that employs circular correlation on entity embeddings and Neural Association Models (NAM)  ~\citep{LiuJiaEvdLinZhuXiaWeiHu16}, a deep network used for probabilistic reasoning.

{\bf Translation Based Models.}
~\citep{BorWesColBen11} uses two relation-specific matrices to project subject and object entities and computes $L_1$ distance to score a fact between two entity vectors. ~\cite{BorUsuGarWesetal13} proposed TransE model that computes score as a distance between relation-specific translations of entity embeddings.~\citep{WanZhaFenChe14} improved TransE by allowing entities to have distributed representations on relation specific hyperplane where distance between them is computed. TransR~\citep{LinLiuSunZhu15} extends this model to use separate semantic spaces for entities and relations and does translation in the relationship space.

\citep{NicMurTreGab16} and ~\citep{YanYihHeGaoDen15,TouChe15} contains comprehensive reviews and empirical comparison of relational learning techniques respectively. All these methods consider knowledge graphs as static models and lack ability to capture temporally evolving dynamics. 

\vspace{-0.2cm}
\subsection{Temporal Modeling}

Temporal point processes have been shown as very effective tool to model various intricate temporal behaviors in networks ~\citep{YanZha13,FarDuGomValZhaSon14,FarWanGomLietal15, DuWanHeetal15, DuDaiTriUpaGomSon16, WanDuTriSon16,  WanTheVerSon16,WanXieDuSon16, WanWilTheSon17, WanYeZhaSon17}. Recently, ~\citep{WanDuTriSon16,DaiWanTriSon16} proposed novel co-evolutionary feature embedding process that captures self-evolution and co-evolution dynamics of users and items interacting in a recommendation system. In relational setting,  
~\citep{LogCecMal15} proposed relational mining approach to discover changes in structure of dynamic network over time. ~\citep{LogMal17} proposes method to capture temporal autocorrelation in data to improve predictive performance.
~\citep{ShaNev08} proposes summarization techniques to model evolving relational-temporal domains. 
Recently,~\citep{EstTreYanBaiSteKro16} proposed multiway neural network architecture for modeling event based relational graph. The authors draw a synergistic relation between a static knowledge graph and an event set wherein the knowledge graph provide information about entities participating in events and new events in turn contribute to enhancement of knowledge graph. They do not capture the evolving dynamics of entities and model time as discrete points which limits its capacity to model complex temporal dynamics. ~\citep{JiaLiuGeLeiLiChaSui16}  models dependence of relationship on time to facilitate time-aware link prediction but do not capture evolving entity dynamics.

\vspace{-0.2cm}
\section{Conclusion}

We propose a novel deep evolutionary knowledge network that efficiently learns non-linearly evolving entity representations over time in multi-relational setting. Evolutionary dynamics of both subject and object entities are captured by deep recurrent architecture that models historical evolution of entity embeddings in a specific relationship space. The occurrence of a fact is then modeled by multivariate point process that captures temporal dependencies across facts. The superior performance and high scalability of our method on large real-world temporal knowledge graphs demonstrate the importance of supporting temporal reasoning in dynamically evolving relational systems.  Our work establishes previously unexplored connection between relational processes and temporal point processes with a potential to open a new direction of research on reasoning over time.

\section*{Acknowledgement}
This project was supported in part by NSF IIS-1218749, NIH BIGDATA 1R01GM108341, NSF CAREER IIS-1350983, NSF IIS-1639792 EAGER, ONR N00014-15-1-2340, NVIDIA, Intel and Amazon AWS.

\vspace{-0.4cm}
\bibliography{bibfile}
\bibliographystyle{icml2017}
\newpage
\onecolumn
\appendix
\section*{Appendix}
\section{Algorithm for Global BPTT Computation}
\label{sec:alg2}
As mentioned in Section 4 of main paper, the intricate relational and temporal dependencies between data points in our setting limits our ability to efficiently train by decomposing events into independent sequences. To address this challenge, we design an efficient Global BPTT algorithm presented below. During each step of training, we build computational graph using consecutive events in the sliding window of a fixed size. We then move sliding window further and train till the end of timeline in similar fashion which allows to capture dependencies across batches while retaining efficiency.

\setcounter{algorithm}{1}
\begin{algorithm}[h!]
   \caption{Global-BPTT}
   \label{alg:alg1}
\begin{algorithmic}
   \STATE {\bfseries Input:} Global Event Sequence $\mathcal{O}$, Steps $s$, Stopping Condition $max\_iter$
   \STATE $cur\_index = 0$, $t\_begin = 0$
   \FOR{$iter=0$ {\bfseries to} $max\_iter$}
   \IF {$cur\_index > 0$}
   \STATE $t\_begin = \mathcal{O}[cur\_index - 1] \rightarrow t$
   \ENDIF
   \STATE e\_mini\_batch = $\mathcal{O}[cur\_index : cur\_index + s]$
   \STATE Build Training Network specific to e\_mini\_batch
   \STATE Feed Forward inputs over network of $s$ time steps
   \STATE Compute Total Loss $\mathcal{L} $ over $s$ steps:
   \STATE $\mathcal{L} = - \sum_{p=1}^s \log\rbr{\lambda^{e^s,e^o}_{r}(t_p|\bar{t_p})}$ + Survival loss computed using Algorithm~\ref{alg:alg2}
   \STATE Backpropagate error through $s$ time steps and update all weights
   \IF {$cur\_index + s > \mathcal{O}.size$}
   \STATE $cur\_index = 0$
   \ELSE 
   \STATE $cur\_index = cur\_index + s$ 
   \ENDIF 

   \ENDFOR
\end{algorithmic}
\end{algorithm}

\newpage
\vspace{-0.3cm}
\section{Data Statistics and Sparsity of Knowledge Tensor}
\label{sec:stats}

\begin{table}[h]
\parbox{.45\textwidth}{
\caption{Statistics for each dataset.}
\centering
\begin{tabular}{cccc}
\toprule
Dataset Name & \# Entities & \# Relations & \# Events\\
&&&\\
\midrule
GDELT-full & 14018 & 20 & 31.29M\\
GDELT-500 & 500 & 20 & 3.42M\\
ICEWS-full & 12498 & 260 & 0.67M\\
ICEWS-500 & 500 & 256 & 0.45M\\
\bottomrule
\label{tab:data_stat}
\end{tabular}
}
\quad\quad
\parbox{.45\textwidth}{
\caption{Sparsity of Knowledge Tensor.}
\centering
\begin{tabular}{cccc}
\toprule
Dataset Name & \# Possible & \# Available & \% Proportion\\
& Entries & Entries & \\
\midrule
GDELT-full & 3.93B & 4.52M & 0.12\\
GDELT-500 & 5M & 0.76M & 15.21\\
ICEWS-full & 39.98B & 0.31M & 7e-3\\
ICEWS-500 & 64M & 0.15M & 0.24\\
\bottomrule
\label{tab:spar_stat}
\end{tabular}
}
\vspace{-0.8cm}
\end{table}

\section{Implementation Details}
\label{sec:impl}
{\bf Know-Evolve.} Both Algorithm~\ref{alg:alg2} and Algorithm~\ref{alg:alg1}  demonstrate that the computational graph for each mini-batch will be significantly different due to high variations in the interactions happening in each window. To facilitate efficient training over dynamic computational graph setting, we leverage on graph embedding framework proposed in~\citep{DaiDaiSon16} that allows to learn over graph structure where the objective function may potentially have different computational graph for each batch. We use Adam Optimizer with gradient clipping for making parameter updates. Using grid search method across hyper-parameters, we set mini-batch size = 200, weight scale = 0.1 and learning rate = 0.0005 for all datasets. We used zero initialization for our entity embeddings which is reasonable choice for dynamically evolving entities. 

{\bf Competitors.} We implemented all the reported baselines in Tensorflow and evaluated all methods uniformly. For each method, we use grid search on hyper-parameters and embedding size and chose the ones providing best performance in respective methods. All the baseline methods are trained  using contrastive max-margin objective function described in~\citep{SocCheManNg13}. We use Adagrad optimization provided in Tensorflow for optimizing this objective function. We randomly initialize entity embeddings as typically done for these models.

\section{Parameter Complexity Analysis}
\label{sec:param_comp}
We report the dimensionality of embeddings and the resulting number of parameters of various models. Table~\ref{tab:param_stat} illustrates that Know-Evolve is significantly efficient in the number of parameters compared to Neural Tensor Network while being highly expressive as demonstrated by its prediction performance in Section 5 of main paper. The overall number of parameters for different dataset configurations are comparable to the simpler relational models in order of magnitude.

\begin{table*}[h!]
\resizebox{\textwidth}{!}{
\begin{tabular}{cccccc}
\toprule
Method & Memory Complexity & \multicolumn{2}{c}{GDELT} & \multicolumn{2}{c}{ICEWS}\\
\cmidrule(l){3-4}\cmidrule(l){5-6}
& & {$H_e/H_r/H_a/H_b$} & {\# Params} & {$H_e/H_r/H_a/H_b$} & {\# Params}\\
\midrule
NTN & $N_e^2H_b + N_r(H_b + H_a) + 2N_rN_eH_a + N_eH_e$ & 100/16/60/60 & 11.83B & 60/32/60/60 & 9.76B\\
RESCAL & $N_rH_e^2 + N_eH_e$ & 100/-/-/- & 1.60M & 60/-/-/- & 1.69M\\
TransE & $N_eH_e + N_rH_e$ & 100/-/-/- & 1.40M & 60/-/-/- & 0.77M\\
TransR & $N_eH_e + N_rH_r + N_rH_r^2$ & 100/20/-/- & 1.41M  & 60/32/-/- & 1.02M\\
ER-MLP & $N_eH_e + N_rH_r + H_a + H_a(2H_e + H_r)$ & 100/20/100/- & 1.42M & 60/32/60/- & 0.77M\\
\midrule
Know-Evolve & $H_e(N_e + N_rH_e) + N_rH_r + H_a*(2H_e + H_r) + H_a*H_b + 2H_b$ & 100/20/100/100 &  1.63M & 60/32/60/60 & 1.71M\\
\bottomrule
\end{tabular}
}
\caption{Comparison of our method with various relational methods for memory complexity. Last two columns provide example realizations of this complexity in full versions for GDELT and ICEWS datasets. $H_a$ and $H_b$ correspond to hidden layers used in respective methods.$H_e$ and $H_r$ correspond to entity and relation embedding dimensions respectively. $N_e$ and $N_r$ are number of entities and relations in each dataset. For GDELT, $N_e = 14018$ and $N_r = 20$. For ICEWS, $N_e = 12498$ and $N_r = 260$. We borrow the notations from~\citep{NicMurTreGab16} for simplicity.}
\label{tab:param_stat}
\end{table*}

\newpage
\section{Exploratory Analysis}
\label{sec:Expl}
\subsection{Temporal Reasoning}

We have shown that our model can achieve high accuracy when predicting a future event triplet or the time of event. Here, we present two case studies to demonstrate the ability of evolutionary knowledge network  to perform superior reasoning across multiple relationships in the knowledge graphs.

\subsubsection*{Case Study I: Enemy's Friends is an Enemy}

\setlength{\intextsep}{-5pt}

\begin{wrapfigure}{r}{0.48\textwidth}
\small
\centering
\includegraphics[width = 0.48\textwidth]{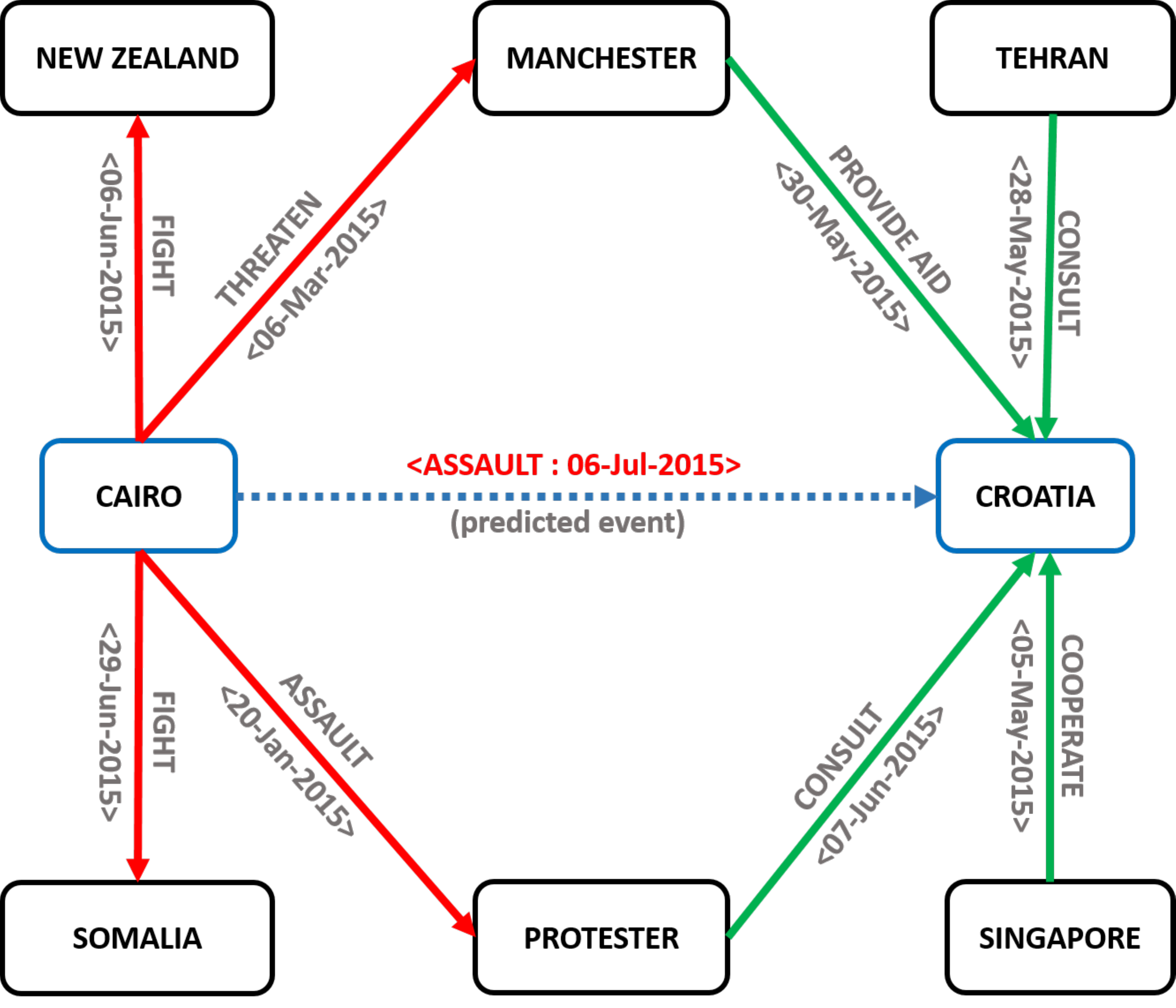}
\caption{Relationship graph for Cairo and Croatia. Dotted arrow shows the predicted edge. Direction of the arrow is from subject to object entity.}
\label{fig:reln1}
\end{wrapfigure}

We concentrate on the prediction of a quadruplet (Cairo,Assault,Croatia,July 5,2015) available in test set. This event relates to the news report of an  assault on a Croation prisoner in Cairo on July 6 2015. Our model gives rank-1 to the object entity \underline{Croatia} while the baselines did not predict them well ($rank > 250$).

We first consider relationship characteristics for Cairo and Croatia. In the current train span, there are $142$ nodes with which Cairo was involved in a relationship as a subject (total of 1369 events) and  Croatia was involved in a relationship as an object (total of 1037 events). As a subject, Cairo was involved in an assault relationship only 59 times while as an object, Croatia was involved in assault only 5 times. As mentioned earlier, there was no direct edge present between Cairo and Croatia with relationship type assault. 

While the conventional reasoning methods consider static interactions of entities in a specific relationship space, they fail to account for the temporal effect on certain relationships and dynamic evolution of entity embeddings. We believe that our method is able to capture this multi-faceted knowledge that helps to reason better than the competitors for the above case. 

{\bf Temporal Effect.} It is observed in the dataset that many entities were involved more in negative relationships in the last month of training data as compared to earlier months of the year. Further, a lot of assault activities on foreign prisoners were being reported in Cairo starting from May 2015. Our model successfully captures this increased intensity of such events in recent past. The interesting observation is that overall, Cairo has been involved in much higher number of positive relationships as compared to negative ones and that would lead conventional baselines to use that path to reason for new entity -- instead our model tries to capture effect of most recent events. 

{\bf Dynamic Knowledge Evolution.} It can be seen from the dataset that Cairo got associated with more and more negative events towards the mid of year 2015 as compared to start of the year where it was mostly involved in positive and cooperation relationships. While this was not very prominent in case of Croatia, it still showed some change in the type of relationships over time. There were multiple instances where Cairo was involved in a negative relationship with a node which in turn had positive relationship with Croatia. This signifies that the features of the two entities were jointly and non-linearly evolving with the features of the third entity in different relationship spaces.

Below we provide reference links for the actual event news related to the edges in Figure~\ref{fig:reln1}.

{\bf Predicted Edge.}
\begin{itemize}
\item (Cairo, Assault, Croatia, 06-Jul-2015): \url{https://www.bloomberg.com/news/articles/2015-08-05/islamic-state-egypt-affiliate-threatens-to-kill-croatian-citizen}
\end{itemize}
{\bf Other Edges.}
\begin{itemize}
\item (Cairo, Assault, Protester, 20-Jan-2015):\url{http://usa.chinadaily.com.cn/world/2015-04/22/content\_20501452}
\item (Cairo, Threaten, Manchester, 06-Mar-2015): \url{http://www.manchestereveningnews.co.uk/news/greater-manchester-news/anthony-filz-stashed-deadly-machine-8788541}
\item (Protester, Consult, Croatia, 07-Jun-2015): 
\url{http://globalvoicesonline.org/2015/06/07/veterans-of-croatias-war-of-independence-are-still-knocking-on-the-governments-door/}
\item (Manchester, Provide Aid, Croatia, 30-May-2015): 
\url{http://www.offthepost.info/blog/2015/05/liverpool-meet-inter-to-discuss-mateo-kovacic-deal/} 
\end{itemize}

\subsubsection*{Case Study II: Common enemy forges friendship}

\setlength{\intextsep}{-20pt}
\begin{wrapfigure}{r}{0.48\textwidth}
\small
\centering
\includegraphics[width = 0.48\textwidth]{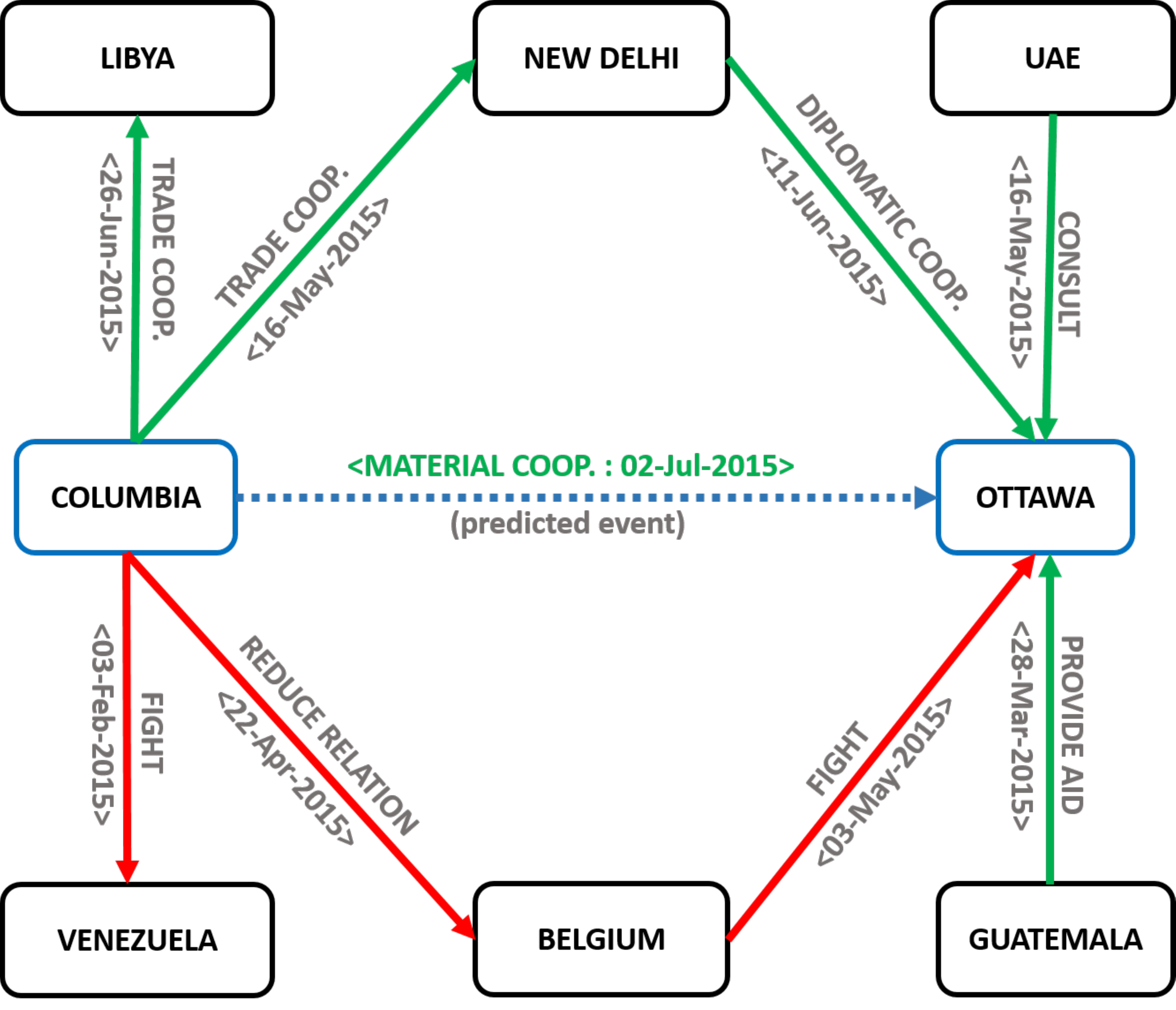}
\caption{Relationship graph for Columbia and Ottawa. Dotted arrow shows the predicted edge. Direction of the arrow is from subject to object entity.}
\label{fig:reln5}
\end{wrapfigure}

We concentrate on the prediction of a quadruplet (Colombia,Engage in Material Cooperation,Ottawa,July 2 2015) available in test set. This event relates to the news report of concerns over a military deal between Colombia and Canada on July 2 2015 and reported in Ottawa Citizen. Our model gives rank-1 to the object entity \underline{Ottawa} while the other baselines do not predict well ($rank > 250$). The above test event is a new relationship and was never seen in training. 

As before, we consider relationship characteristics between Colombia and Ottawa. In the current train span, there are $165$ nodes for which Colombia was involved in a relationship with that node as a subject (total of 1604 events) and on the other hand, Ottawa was involved in a relationship with those nodes as an object total of 733 events). As a subject, Colombia was involved in a cooperation relationship 71 times while as an object, Ottawa was involved in cooperation 24 times. \\

{\bf Temporal Effect.} It is observed in the dataset that Colombia has been involved in hundreds of relationships with Venezuela (which is natural as they are neighbors). These relationships range across the spectrum from being as negative as ``fight" to being as positive as ``engagement in material cooperation". But more recently in the training set (i.e after May 2015), the two countries have been mostly involved in positive relationships. Venezuela in turn has only been in cooperation relationship with Ottawa (Canada). Thus, it can be inferred that Colombia is affected by its more recent interaction with its neighbors while forming relationship with Canada.

{\bf Dynamic Knowledge Evolution.} Overall it was observed that Colombia got involved in more positive relationships towards the end of training period as compared to the start. This can be attributed to events like economic growth, better living standards, better relations getting developed which has led to evolution of Colombia's features in positive direction. The features for Ottawa (Canada) have continued to evolve in positive direction as it has been involved very less in negative relationships. 

More interesting events exemplifying mutual evolution were also observed. In these cases, the relationship between Colombia and third entity were negative but following that relationship in time, the third entity forged a positive relationship with Ottawa (Canada). One can infer that it was in Colombia's strategic interest to forge cooperation (positive relation) with Ottawa so as to counter its relationship with third entity. 
%
Below we provide reference links for the actual event news related to the edges in Figure~\ref{fig:reln5}.

{\bf Predicted Edge.}
\begin{itemize}
\item (Columbia, Material Coop., Ottawa, 02-Jul-2015): \url{http://ottawacitizen.com/news/politics/report-on-military-executions-casts-shadow-over-lav-deal-with-colombia}\\\\
\end{itemize}
{\bf Other Edges.}
\begin{itemize}
\item (Columbia, Trade Coop., New Delhi, 16-May-2015): \url{http://www.newindianexpress.com/business/2015/may/16/Petroleum-Minister-Dharmendra-to-Lead-Business-Delegation-to-Mexico-Colombia-761494.html}
\item (Columbia, Fight, Venezuela, 03-Feb-2015):\url{http://www.turkishpress.com/news/421947/}
\item (New Delhi, Diplomatic Coop., Ottawa, 28-May-2015):\url{http://www.marketwatch.com/story/art-of-living-set-to-showcase-the-yoga-way-2015-06-11-61734555}
\item (Belgium, Fight, Ottawa, 05-May-2015): \url{https://www.durhamregion.com/news-story/5597504-9-facts-about-in-flanders-fields-on-its-100th-anniversary/}
\end{itemize}

\setlength{\intextsep}{5pt}
\subsection{Sliding Window Training Experiment} 
\begin{figure}[h!]
\small
\centering
\resizebox{0.50\textwidth}{!}{
\begin{tabular}{ccc}
\rotatebox{90}{\bf\small~~~~~~~{GDELT-500}}~~
&\includegraphics[width = 0.26\textwidth]{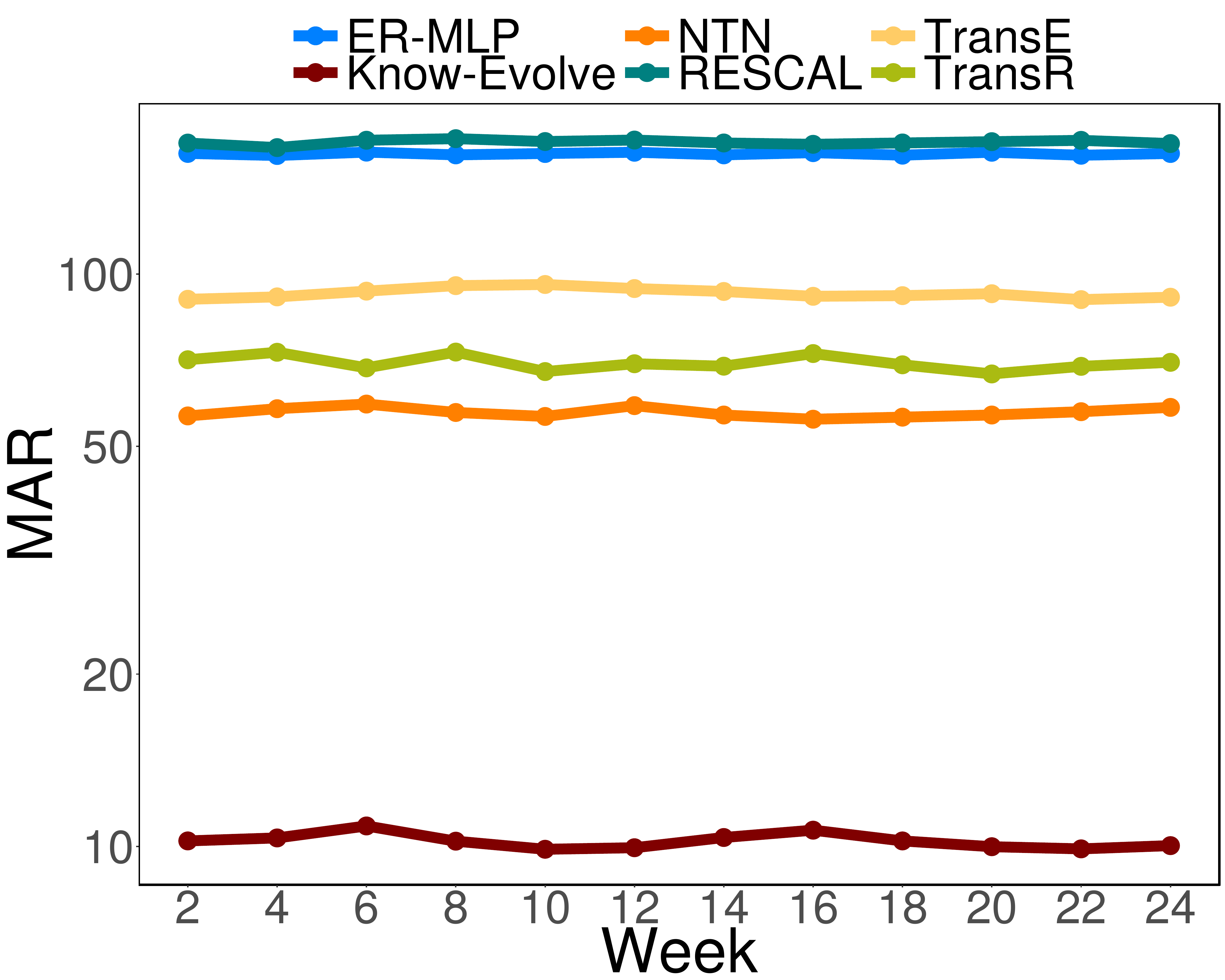}
& \includegraphics[width = 0.26\textwidth]{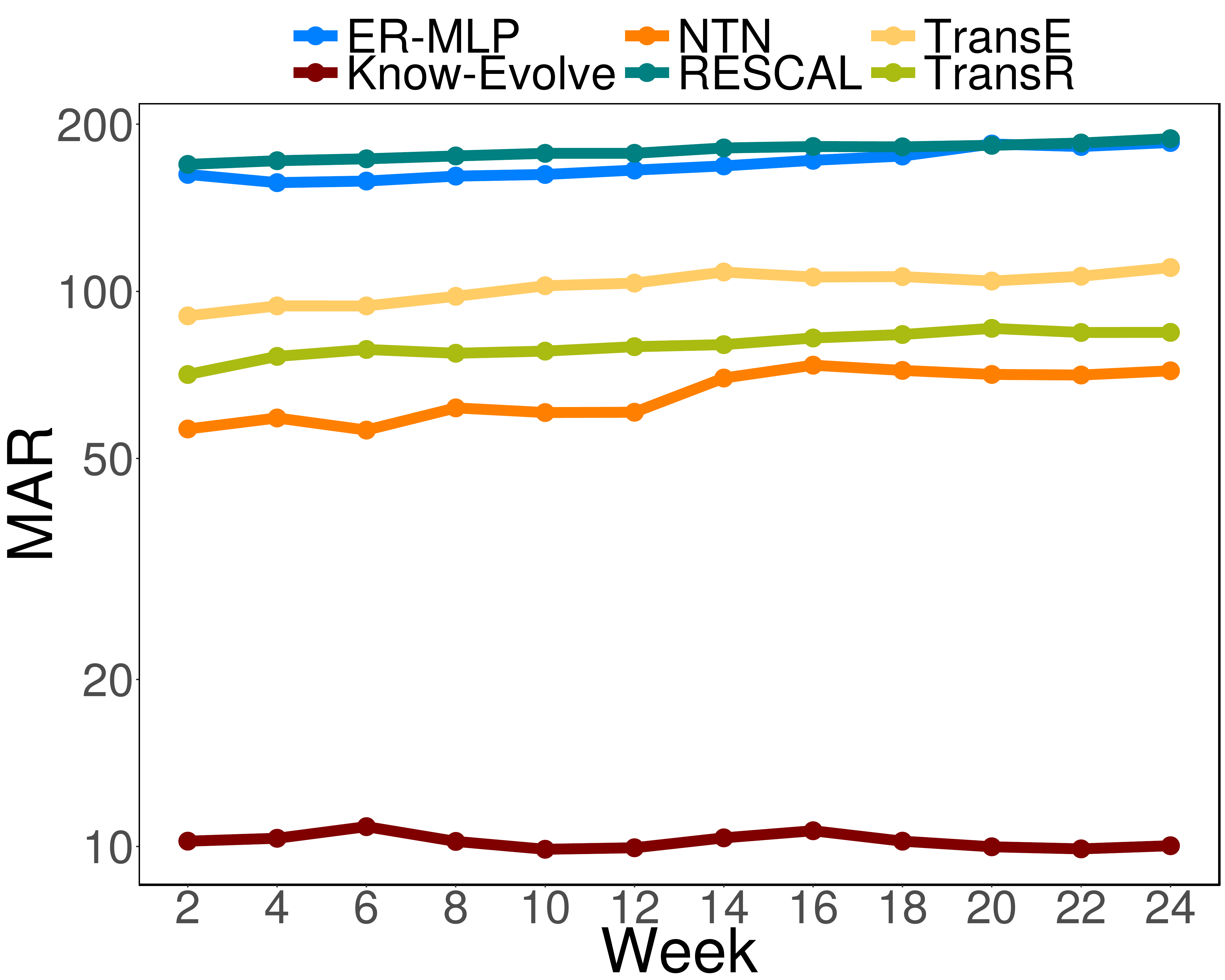}\\
& (c) Sliding Window Training & (d) Non-sliding window Training 
\end{tabular}
}
\vspace{-2mm}
\caption{Performance comparison of sliding window vs. non-sliding window training (in terms of link prediction rank).}
\label{fig:comp}
\vspace{-2mm}
\end{figure}

Unlike competitors, the entity embeddings in our model get updated after every event in the test, but the model parameters remain unchanged after training. To balance out the advantage that this may give to our method, we explore the use of sliding window training paradigm for baselines: We train on first six months of dataset and evaluate on the first test window. Next we throw away as many days (2 weeks) from start of train set as found in test set and incorporate the test data into training. We retrain the model using previously
learned parameters as warm start. This can effectively aid the baselines to adapt to the evolving knowledge over time. Figure~\ref{fig:comp} shows that the sliding window training contributes to stable performance of baselines across the time window (i.e.the temporal deterioration is no longer observed significantly for baselines). But the overall performance of our method still surpasses all the competitors.

\subsection{Recurrent Facts vs. New facts} 
\begin{figure}[h!]
\small
\centering
\resizebox{0.50\textwidth}{!}{
\begin{tabular}{cc}
 \includegraphics[width = 0.25\textwidth]{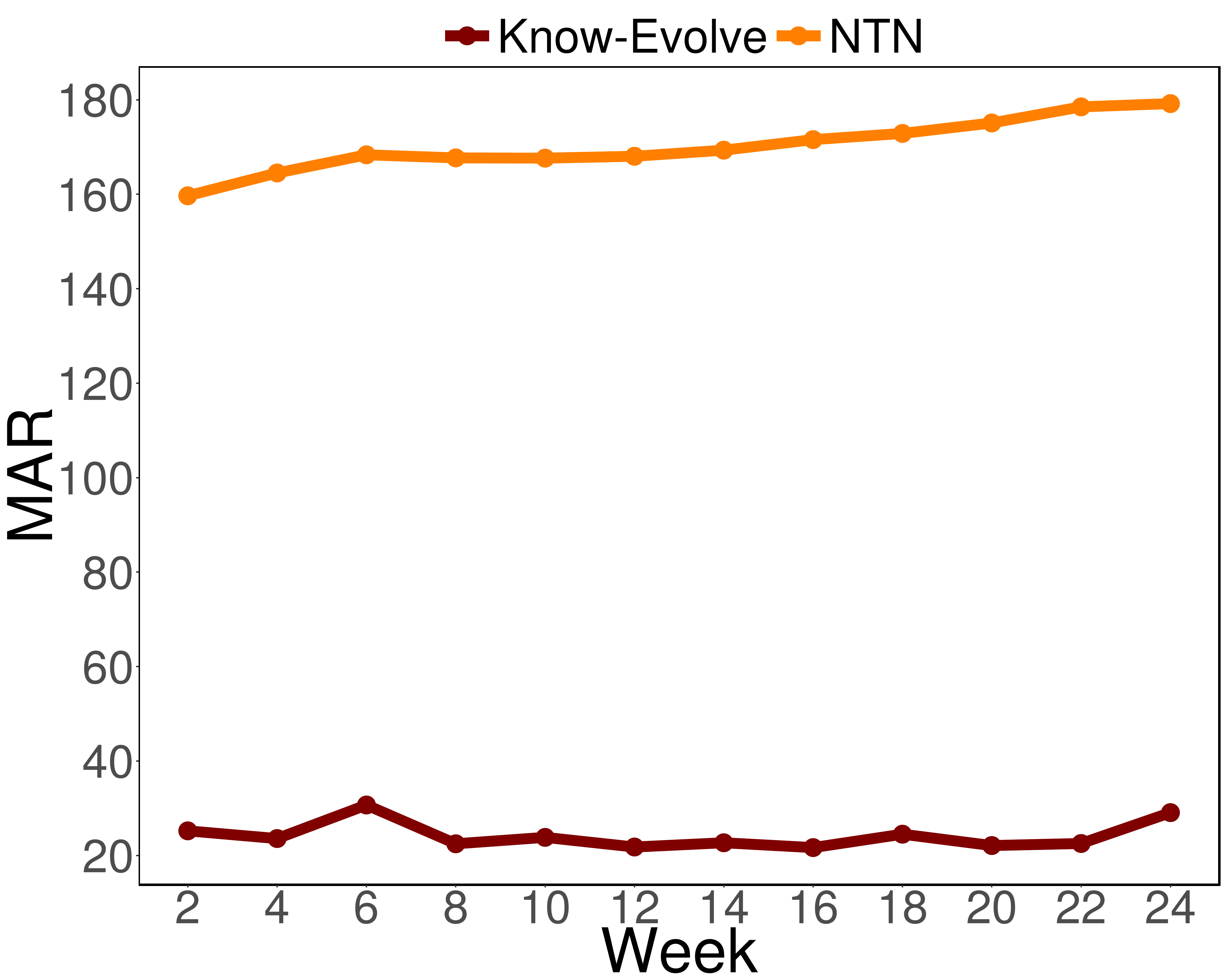}
& \includegraphics[width = 0.25\textwidth]{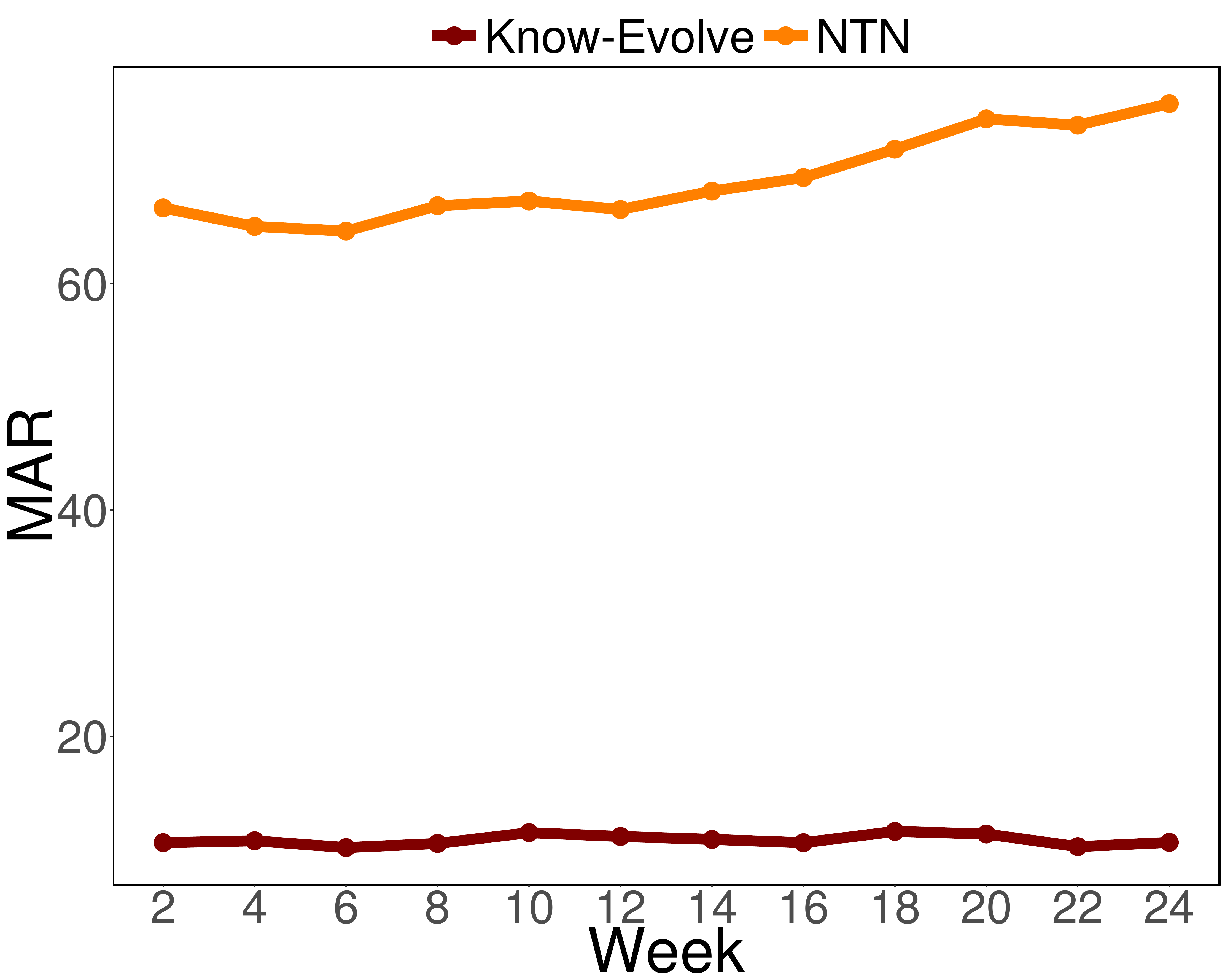}\\
(a) New facts only & (b) Recurrent Facts Only
\end{tabular}
}
\vspace{-2mm}
\caption{Comparison with NTN over recurrent and non-recurrent test version.}
\label{fig:rec_pred}
\end{figure}

One fundamental distinction in our multi-relational setting is the existence of recurrence relations which is not the case for traditional knowledge graphs. To that end, we compare our method  with the best performing competitor - NTN on two different testing setups: 1.) Only Recurrent Facts in test set 2.) Only New facts in test set. We perform this experiment on GDELT-500 data. We call a test fact ``new" if it was never seen in training. As one can expect, the proportion of new facts will increase as we move further in time. In our case, it ranges from 40\%-60\% of the total number of events in a specific test window. Figure \ref{fig:rec_pred} demonstrates that our method performs consistently and significantly better in both cases.

\end{document}